\documentclass[letterpaper]{article} 
\usepackage{aaai24}  
\usepackage{times}  
\usepackage{helvet}  
\usepackage{courier}  
\usepackage[hyphens]{url}  
\usepackage{graphicx} 
\usepackage{natbib}  
\usepackage{caption} 
\usepackage{algorithm}
\usepackage{algorithmic}
\usepackage{amsfonts}

\usepackage{amsmath}
\usepackage{amssymb}
\usepackage{mathtools}
\usepackage{amsthm}
\usepackage{cuted}
\usepackage{booktabs}
\usepackage{tabularray}
\usepackage{arydshln}
\usepackage{caption}
\usepackage{subcaption}
\usepackage{multirow}

\usepackage{enumitem,amssymb}
\newlist{todolist}{itemize}{2}
\setlist[todolist]{label=$\square$}
\usepackage{pifont}
\newcommand{\xmark}{\ding{55}}%

\theoremstyle{plain}
\newtheorem{theorem}{Theorem}[section]
\newtheorem{proposition}[theorem]{Proposition}

\newtheorem{corollary}[theorem]{Corollary}
\theoremstyle{definition}
\newtheorem{definition}[theorem]{Definition}
\newtheorem{assumption}[theorem]{Assumption}
\theoremstyle{remark}

\usepackage{soul} 
\usepackage{url}
\usepackage{subfiles}
\DeclareMathOperator{\always}{\square}
\DeclareMathOperator{\nextltl}{\bigcirc}
\DeclareMathOperator{\eventually}{\Diamond}
\DeclareMathOperator{\until}{\mathbf{U}}

\usepackage{newfloat}
\usepackage{listings}
\DeclareCaptionStyle{ruled}{labelfont=normalfont,labelsep=colon,strut=off} 
\floatstyle{ruled}
\newfloat{listing}{tb}{lst}{}
\floatname{listing}{Listing}
\title{A Reminder of its Brittleness: Language Reward Shaping May Hinder Learning for Instruction Following Agents}
\author {
    Sukai Huang,
    Nir Lipovetzky and 
    Trevor Cohn\thanks{Now at Google DeepMind}
}
\affiliations {
    School of Computing and Information Systems, The University of Melbourne, Australia\\
    sukaih@student.unimelb.edu.au, \{nir.lipovetzky,
    trevor.cohn\}@unimelb.edu.au \\
}

\begin{document}

\maketitle

\begin{abstract}
Teaching agents to follow complex written instructions has been an important yet elusive goal. One technique for enhancing learning efficiency is language reward shaping (LRS). Within a reinforcement learning (RL) framework, LRS involves training a reward function that rewards behaviours precisely aligned with given language instructions. We argue that the apparent success of LRS is brittle, and prior positive findings can be attributed to weak RL baselines. Specifically, we identified suboptimal LRS designs that reward partially matched trajectories, and we characterised a novel reward perturbation to capture this issue using the concept of loosening task constraints. We provided theoretical and empirical evidence that agents trained using LRS rewards converge more slowly compared to pure RL agents. Our work highlights the brittleness of existing LRS methods, which has been overlooked in the previous studies. \footnote{Code and replication materials is released at Github \url{https://github.com/sino-huang/brittleness_of_lrs}.}
\end{abstract}

\section{Introduction}
In recent years, large-scale language models have shown exceptional performance in downstream tasks \citep{fei2022towards, bommasani2021opportunities, duan2022multi,  alayrac2022flamingo}, leading to increased interests in integrating these pretrained language models with task-specific control policies for instruction following agents \citep{luketina2019survey}. Previous literature on instruction following often proposed end-to-end models that directly mapped language instructions and state observations to task-specific low-level control. However, this tight coupling with the task environment prevented the use of general pretrained models.

\begin{figure}[t]
    \centering
    \includegraphics[width=0.8\columnwidth]{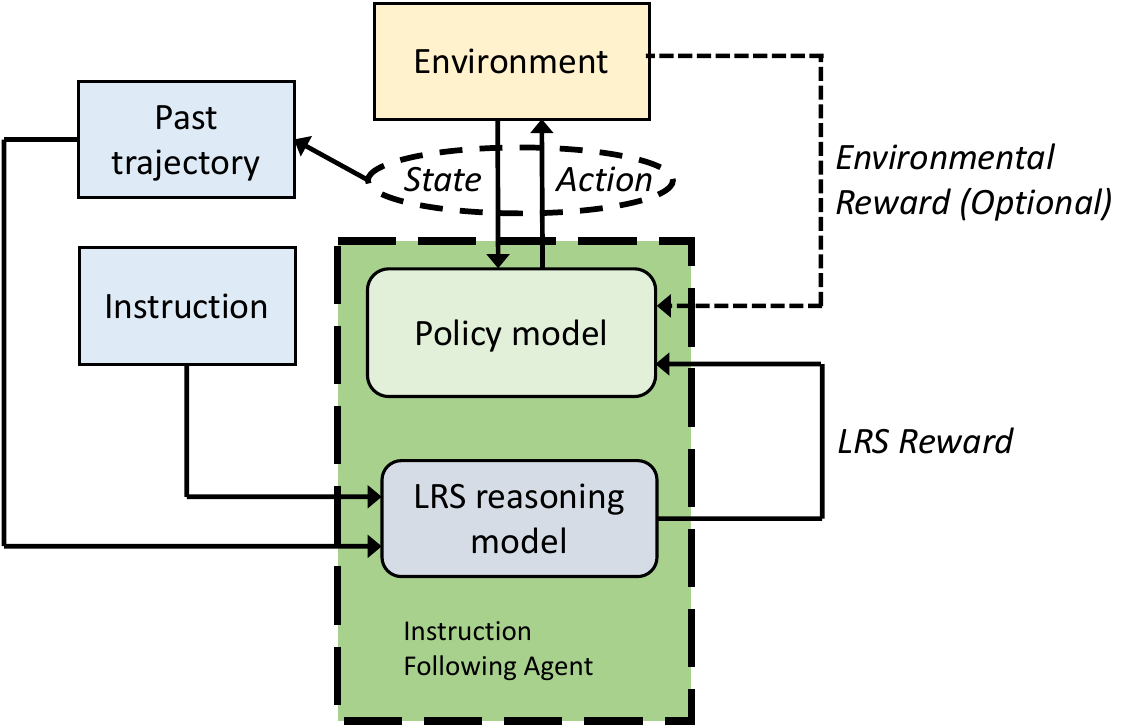}
    \caption{A Language Reward Shaping framework}
    \label{fig:lrsframework}
\end{figure}

To address this limitation, attempts have been made to develop instruction following agents featuring separate reasoning and policy models, allowing pretrained models to be seamlessly integrated. One such methodology is Language Reward Shaping (LRS), which leverages pretrained multimodal or unimodal models to formulate reward functions rewarding RL agents' behaviors in accordance with expert instructions. (see Figure~\ref{fig:lrsframework}). From the perspective of RL researchers, LRS offers a means of tackling the sparse reward challenge by introducing supplementary reward signals through language information. 

The core mechanism of LRS involves compressing high-dimensional semantic vectors of natural language instructions encoded by pretrained large models into scalar rewards. However, despite its apparent simplicity, we have observed that agents trained with LRS rewards often exhibit slower convergence compared to purely RL-trained agents in practical scenarios. Thus, the key research question that we aim to answer in this paper is: \emph{what causes the deterioration of LRS methods?} 

\begin{figure}
    \centering
    \includegraphics[width=0.9\columnwidth]
    {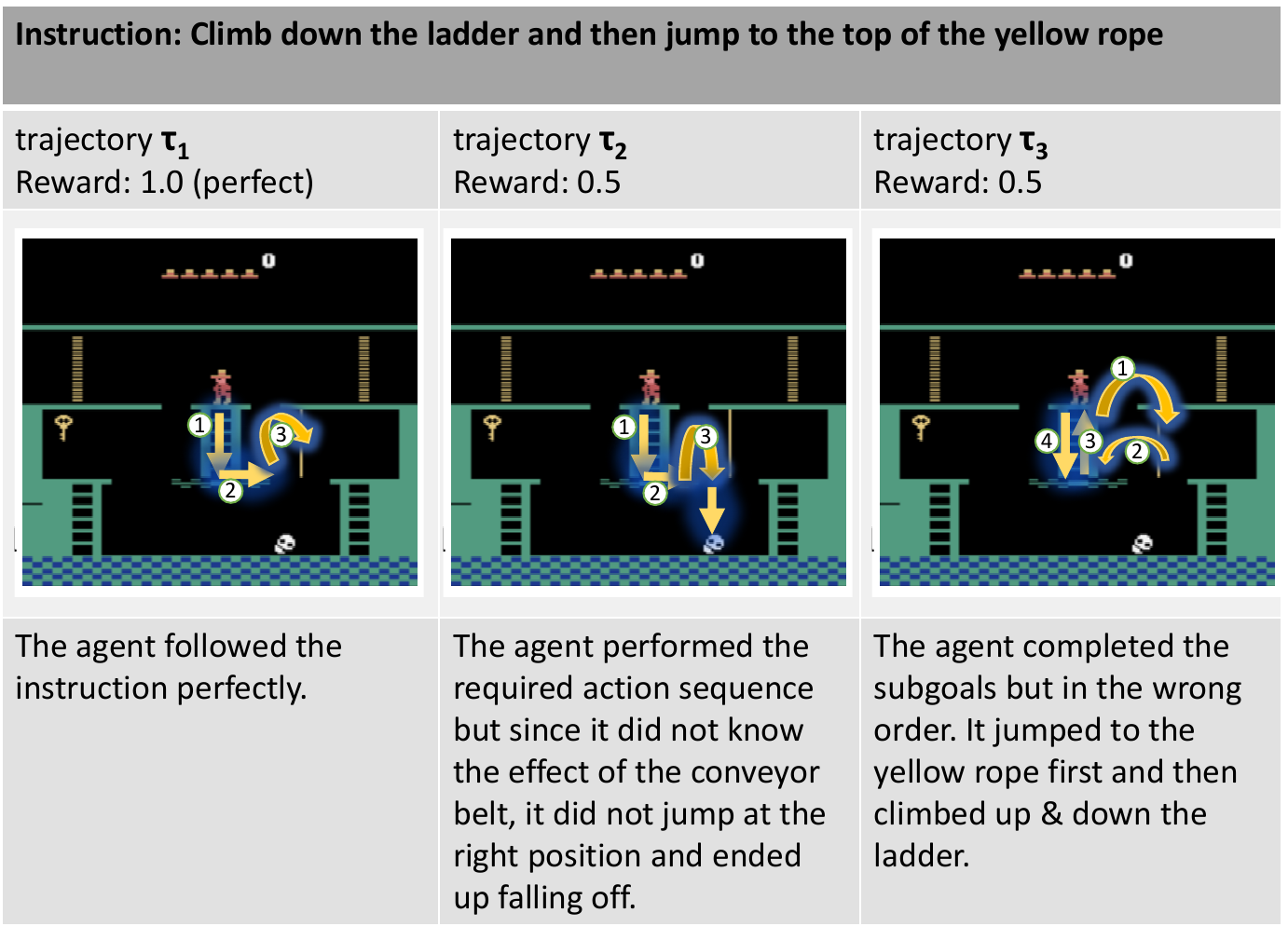}
    \caption{In practice, partially matched trajectories can still be assigned rewards that are reduced but not entirely void.}
    \label{fig:lrsfalsepositiveexample}
\end{figure}

Our contributions in this paper are summarised as follows: 1) we identify suboptimal LRS architecture designs that reward partially matched trajectories (see Figure~\ref{fig:lrsfalsepositiveexample}); 2) we introduce a novel form of reward perturbation that characterizes this issue through the concept of relaxing task constraints; 3) we offer both theoretical and empirical evidence demonstrating that, compared to a reinforcement learning algorithm that effectively explores the state-space, LRS can underperform due to its tendency to reward partially matched trajectories. However, it is important to clarify that we don't seek to undermine the value of LRS methodology. Rather, our goal is to offer a perspective on potential factors contributing to slow convergence.

\section{Related Work}
\label{sec:relatedwork}

\textbf{Language Reward Shaping (LRS)} is a term formally defined by \citet{goyal2019using, goyal2021pixl2r}. It involves constructing a reward function that rewards agent behaviours aligned with expert instructions. The technique has been implemented across various domains, ranging from dialogue systems \citep{ouyang2022training}, visual language navigation \citep{wang2019reinforced} to recommender systems \citep{lin2022inferring}. While prior studies on LRS \citep{kaplan2017beating, ibarz2018reward, wang2019reinforced, du2023guiding} have touted its efficacy, they have not discussed the potential collapse of the system due to rewarding partially matched trajectories. 

\noindent \textbf{Reward Perturbation Analysis} is an orthogonal work that studies the influence of reward perturbation on the performance of RL algorithms. Previous research on reward perturbation has focused on relatively limited types of poisoned rewards, such as reversing the sign of the numeric rewards \citep{zhang2021robust, ilahi2021challenges}, adaptive reward attack \citep{huang2017adversarial, zhang2020adaptive}, reward delay attack \citep{sarkar2022reward}, and action-triggered reward manipulation \citep{wang2021backdoorl}. In comparison, our research introduces and evaluates the effects of a new variant of reward perturbation, rooted in the concept of relaxing task constraints, which is implicitly inherent in existing LRS methods.

\section{Problem Setup}
\begin{figure*}[t]
    \centering
     \begin{subfigure}[b]{0.38\textwidth}
         \includegraphics[width=\textwidth]{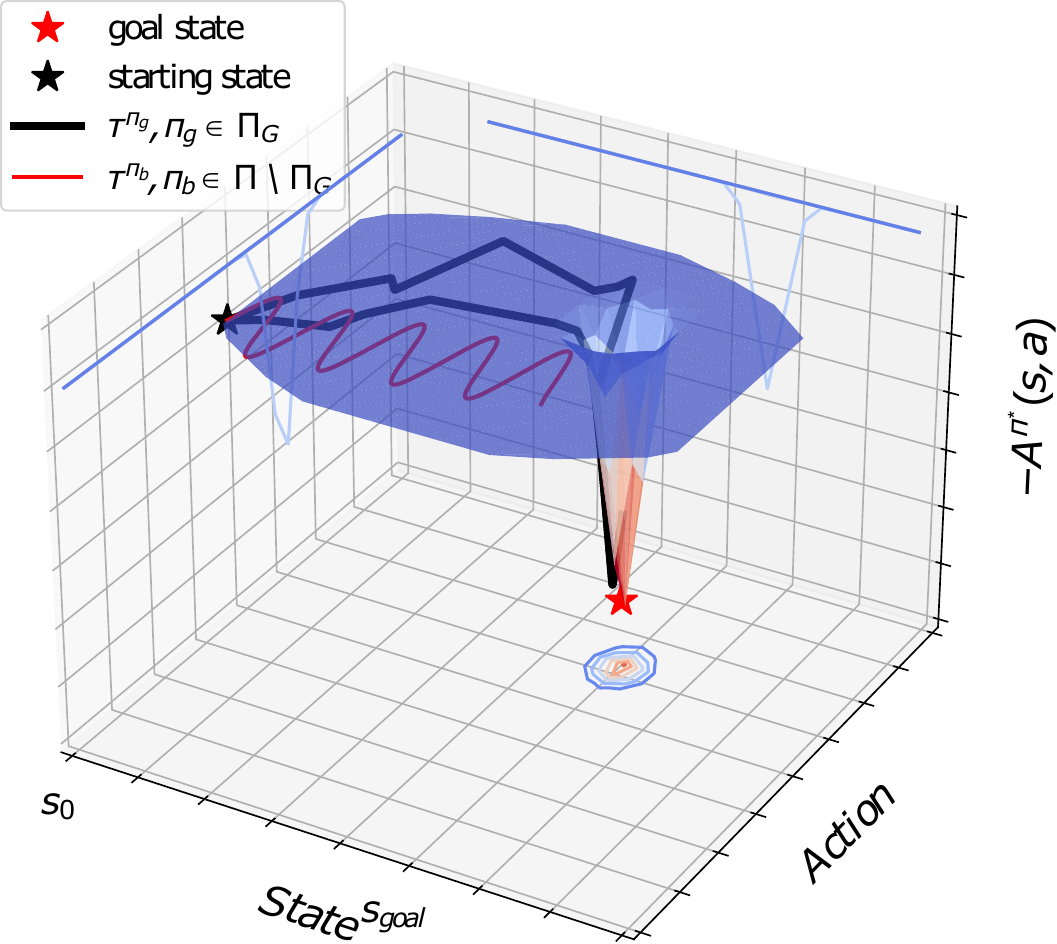}
         \caption{Sparse reward task without LRS}
         \label{fig:sparserewardtaskexample}
     \end{subfigure}
     \hfill
     \begin{subfigure}[b]{0.38\textwidth}
         \includegraphics[width=\textwidth]{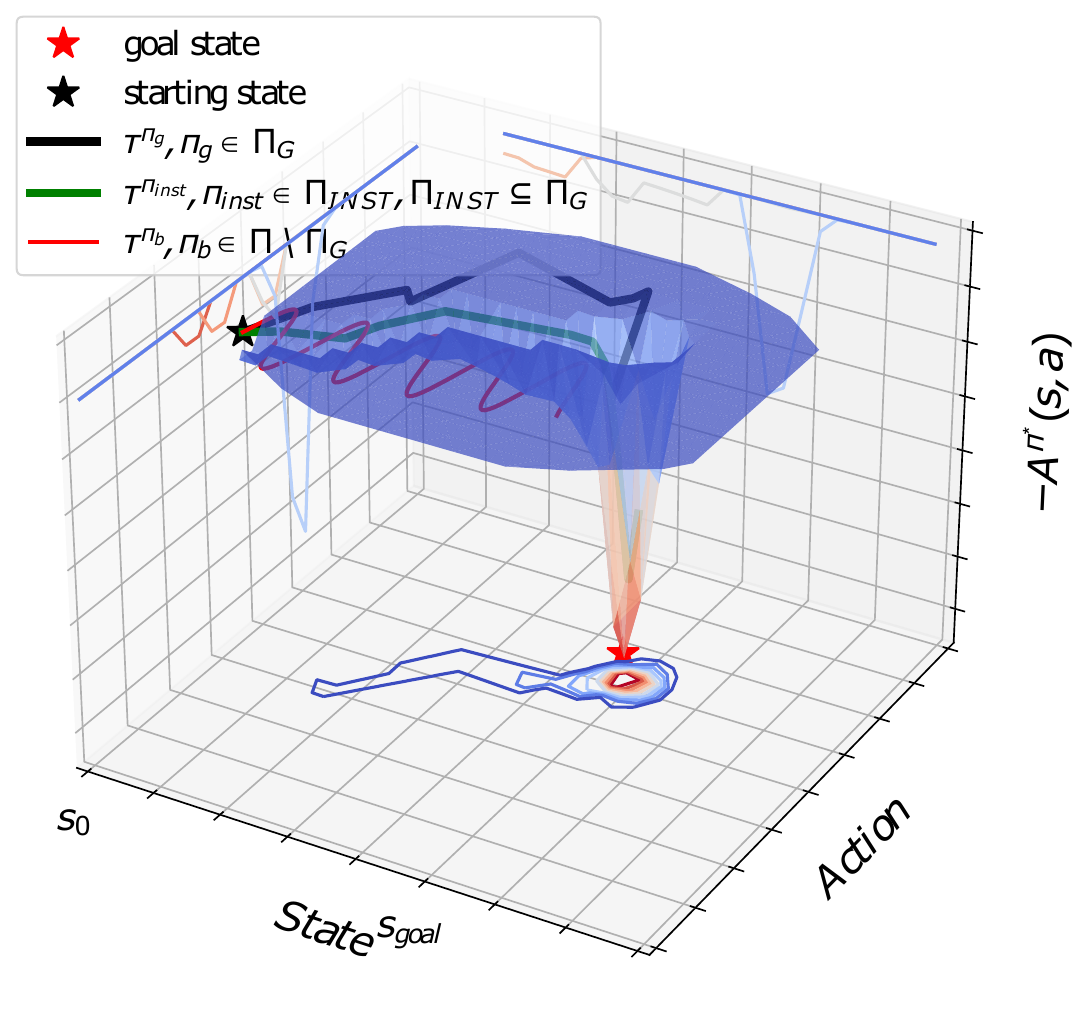}
         \caption{Sparse reward task with LRS}
         \label{fig:sparserewardtaskwithlrsrewards}
     \end{subfigure}
    \caption{A schematic drawing of how LRS helps RL algorithms by providing a gradient on the value function manifold towards the goal state. In a sparse reward environment, finding an acceptable policy $\pi_g \in \Pi_G$ relies heavily on the exploration ability of the agent. In contrast, LRS provides auxiliary rewards to set $\Pi_{\mathit{INST}}$, creating a gradient towards the goal state.}
    \label{fig:sparserewardtaskwandwolrs}
\end{figure*}
\subsection{Sparse Reward MDP Environment}

In this work, we define our task as a MDP tuple $\langle S, A, T, s_0, S_G, S_L, R^{\mathit{env}} \rangle$. Here, $S$ is the finite set of states, $A$ is the finite set of actions, and $T : S \times A \rightarrow S$ is a deterministic state transition function where, for any state $s \in S$ and action $a \in A$, the next state $s'$ is given by $T(s,a)$. The process starts at the unique initial state $s_0$. $S_G$ is a set of goal states, and $S_L$ is a set of terminating states in which every action causes a deterministic transition back to itself with zero reward. The sparse environmental reward function is given by $R^{\mathit{env}} : S \times A \times S \rightarrow \mathbb{R}$, where for any $s, a, s'$:
    \begin{equation*}
    R^{\mathit{env}}(s,a,s') = 
        \begin{cases}
        r & \text{if $s'$ in $S_G$} \\
        0 & \text{otherwise}
        \end{cases}
    \end{equation*}
    where $r$ is a positive constant representing the reward for reaching a goal state.

In addition, we adopted the subsequent assumption:

\begin{assumption}[Deterministic Policies]
We assume deterministic policies $\pi : S \rightarrow A$, which map each state to a specific action.
\end{assumption}

\begin{corollary}[Bijective Correspondence]
Due to the deterministic nature of the transition function $T$ and policies $\pi$, there exists a bijective correspondence between a policy $\pi$ and its trajectory $\tau^\pi$. A trajectory $\tau^\pi$ from start state $s_0$ to a terminating state is a sequence of states and actions $(s_0, a_0, s_1, a_1, ...)$ such that $s_{i+1} = T(s_i, a_i)$ where $a_i = \pi(s_i)$.
\end{corollary}

Adhering to the problem formulation presented by \citet{abel2021expressivity}, we further define: 
\begin{itemize}
    \item \emph{Policy Universe, $\Pi$:} The set of all possible policies for the MDP.
    \item \emph{Acceptable Policies, $\Pi_G$:} A subset of $\Pi$ such that any policy $\pi \in \Pi_G$ has its trajectory $\tau^\pi$ terminating in a goal state $s_g \in S_G$. Formally, $\Pi_G = \{\pi \in \Pi | \exists s_g \in S_G : s_g \text{ is reachable from $s_0$ under $\pi$}\}$
\end{itemize}
This configuration leads to the following proposition: 
\begin{proposition}[\citet{abel2021expressivity}]
    The sparse environmental reward function $R^\mathit{env}$ can effectively distinguish policies from $\Pi_G$ against the complete policy universe $\Pi$, but cannot make distinctions of preference within $\Pi_G$.
\end{proposition}

\subsection{Formulation of Expert Instructions}
Drawing on the principle that expert instructions capture preferences for a subset of acceptable policies, we model the alignment of a trajectory to an expert instruction as a constraint matching problem. In this framework, we conceptualize the instruction as a composite of action, state, and temporal constraints:

\begin{definition}[Constraint Composition]
Let's consider expert instructions that reflect preferences for a subset of acceptable policies, denoted by $\Pi_\mathit{INST} \subseteq \Pi_G$. In this context, we defined that:
\begin{enumerate}
    \item An \emph{action constraint}, denoted by $C_a$, is derived from the action observed in the preferred trajectories taken by acceptable policies. Formally, $C_a := \{a | a \in \tau^\pi, \pi \in \Pi_\mathit{INST} \subseteq \Pi_G, a \in A\}$.
    \item A \emph{state constraint}, denote by $C_s$, is based on states traversed in the preferred trajectories of acceptable policies. Formally, $C_s := \{s | s\in \tau^\pi, \pi \in \Pi_\mathit{INST} \subseteq \Pi_G, s \in S\}$.
    \item A \emph{temporal constraints}, $C_t$, captures the sequential relationship among transitions $T(s_i, a_i)$ in a trajectory. There exists a myriad of formal languages that can express temporal relationships (e.g., Linear Temporal Logic (LTL) \citep{camacho2019ltl}). More detailed information is in Appendix~\ref{sec:appendixLTL}.
\end{enumerate}
\end{definition}

Based on the definition provided above, a trajectory aligns with an expert instruction if and only if it satisfies the associated action, state, and temporal constraints. If the trajectory meets only some of the constraints, we term it as being ``partially matched'' with the instruction.

The choice of formulation has its roots in the compositional nature of natural language. Research indicates that leveraging the compositionality of language enhances problem-solving capacities in language models \citep{akyurek2022compositionality, bhambri2023multi, drexler2021expressing, liu2022planning}. By situating our instruction-following problem within a constraint satisfaction context, we push for a more structured interpretation of language instructions, aligning with insights from \citet{yang2021safe}.

\subsection{Shaping Reward Function by Instructions}

Language Reward Shaping aims to provide reward signals for RL agents when their trajectories match expert preferences encoded in natural language instructions, as depicted in Figure~\ref{fig:sparserewardtaskwandwolrs}. Mathematically, LRS constructs a potential-based shaping reward function $F$ to provide immediate auxiliary rewards for RL algorithms, with convergence guaranteed in theory. 

\begin{proposition}[Convergence guarantee]
\label{prop:conver}
A reward shaping function that gives rewards when agent's trajectory matches the language instruction can be written in the form 
\begin{equation}
\label{eq:potential-based-func-history}
F_t = F(s_t, a_t, s_{t-1}, a_{t-1}) = \Phi(s_t, a_t) - \gamma^{-1} \Phi(s_{t-1}, a_{t-1}) \nonumber
\end{equation}
where $a_t$ is the action for the current state $s_t$ at time $t$; $\Phi : S \times A \rightarrow \mathbb R$ is a real-valued function. Then, the optimal policy $\pi^* \in \Pi_G$ in the original task environment setting will remain optimal in the new setting where the reward signal at time $t$ becomes $R^{\text{env}}_t + F_t$
\end{proposition}
While the shaping reward function, as described in Equation~\ref{eq:potential-based-func-history}, appears \emph{non-Markovian} due to its reliance on how past behavior aligns with the instruction, it's pivotal to understand that the policy's training remains unaffected provided it is an on-policy model. A detailed explanation of this behavior and its implications can be found in Appendix~\ref{sec:appendixproofs}. We say that this form of reward function is potential-based \citep{ng1999policy, wiewiora2003principled}. While theoretical indications suggest that a potential-based LRS function can enhance the learning efficiency of RL algorithms, our investigation unveils that practical convergence can be notably hindered by the issue of rewarding partially matched trajectories.

\section{The Brittleness of LRS}
\label{sec:brittleoflrs}

In this section, we will particularly focus on the specific concern of rewarding partially matched trajectories and provide theoretical evidence demonstrating its impact on the convergence rate of RL algorithms. To illustrate this, we will use Actor Critic as an example.

\begin{figure*}[t]
    \centering
     \begin{subfigure}[b]{0.42\textwidth}
         \centering
         \includegraphics[width=\textwidth]{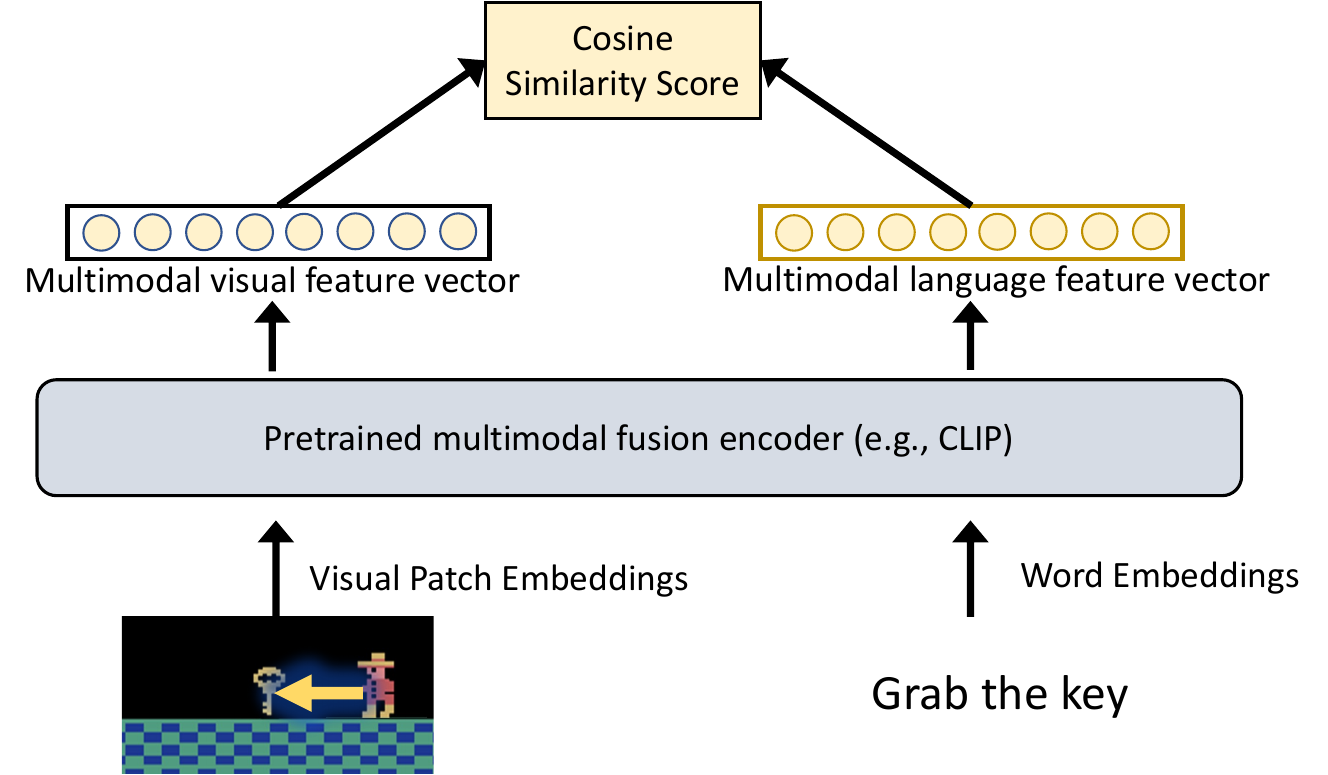}
         \caption{LRS with a cosine similarity representation layer}
         \label{fig:twotypeoflrs_a}
     \end{subfigure}
     \hfill
     \begin{subfigure}[b]{0.42\textwidth}
         \centering
         \includegraphics[width=\textwidth]{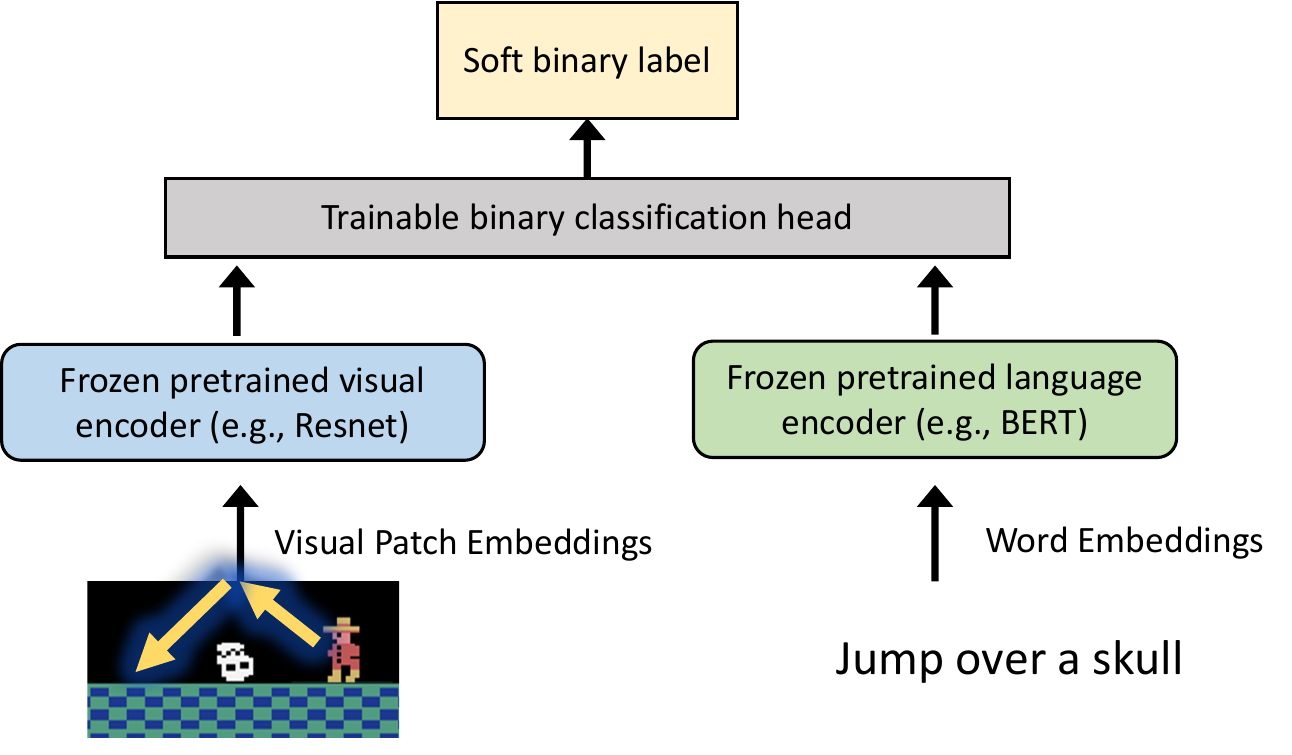}
         \caption{LRS with a trainable binary classifier}
         \label{fig:twotypeoflrs_b}
     \end{subfigure}

    \caption{An illustration of two suboptimal design classes in Language Reward Shaping models. Both design classes fail to avoid rewarding partially matched trajectories.}
    \label{fig:twotypeoflrs}
\end{figure*}

\subsection{Suboptimal Compression of the Information -- the Bottleneck in LRS Models}

LRS operates by compressing high-dimensional semantic vectors into scalar reward signals. However, this compression introduces ambiguity and imprecision about what should be rewarded.

One crucial aspect contributing to the to limitation of LRS architecture is the \emph{smoothness of the model}. As illustrated in Figure~\ref{fig:twotypeoflrs}, LRS models predominantly employ gradient-based learning methods. Among prevalent LRS implementations, two primary prediction styles emerge for matching instruction-trajectory pairs: cosine similarity and binary classification. Works by \citet{kaplan2017beating, kant2022housekeep, du2023guiding} have harnessed cosine similarity in their output layers. In contrast, binary classification output layers are used in works like \citet{goyal2019using} and \citet{wang2019reinforced}. The inherent characteristic of gradient-based learning to allocate non-zero rewards, even for imperfect input alignments, is a double-edged sword. While it leads to a smoother and more exploratory learning process, it concurrently increases the likelihood of agents exploiting towards suboptimal paths. It gets more challenging when attempting to reinforce specific behaviors within a trajectory that were intended to be pivotal in achieving a particular reward. 

Another challenge is the \emph{handling of temporal information}. Both the game environment and the language instructions involve temporal aspects, but Transformer models \citep{vaswani2017attention}, which are commonly used as the backbone architecture in LRS models, may not adequately capture chronological ordering \citep{pham2020out}. While Transformer models can learn temporal dependencies to some extent, there is little in the pre-training to enforce this strictly. Thus, the reward function can be less sensitive to the ordering of action execution. Additionally, the temporal dimension collapses when compressing instructions into scalars. These aforementioned challenges ultimately contribute to the concrete issue of \emph{rewarding partially matched trajectories}.

\subsection{Rewarding Partially Matched Trajectories}

Rewarding partially matched trajectories leads to challenges for learning agents in distinguishing between preferred and non-preferred behavior segments. As shown in the examples in Figure~\ref{fig:lrsfalsepositiveexample}, these false positive rewards might misguide LRS agents into thinking they're on the right track, leading to the repetition of suboptimal behaviors. 

Adding to this complexity is the \emph{compositionality of natural language}. While a finer partitioning of the training dataset may aid in distinguishing between matched and unmatched pairs, an atomic sentence that cannot be further broken down at the natural language level can still encompass multiple constraints. For instance, the atomic sentence ``cut down the tree'' contains both the action constraint ``cutting down'' and the object constraint ``tree''. Thus, even with meticulous annotation, assigning a clear-cut binary label (0 or 1) to instruction and trajectory pairs is not straightforward. It is due to the possibility of trajectories partially aligning with an atomic instruction, thereby blurring the distinction between a complete match and a mismatch. Therefore, in practice, it is infeasible to establish a uniform threshold for distinguishing whether a reward reduction is due to a hard negative or the estimation error.


\subsection{Reduction of Convergence Rate}
\label{sec:convfailure}
Convergence rate refers to the number of training iterations needed to learn an $\epsilon$-optimal policy. However, the exact convergence rate calculations for RL algorithms provided in prior research \citep{agarwal2021theory,xiao2022convergence} cannot be directly applied in this context. This limitation arises from their reliance on the assumption that the initial state is sampled from a certain distribution. Therefore, an alternative way of analysing the convergence rate is to measure the probability of picking acceptable policies (i.e., $P(\pi \in \Pi_G)$) over each update iteration. 

\begin{theorem}[Convergence rate reduction]
\label{coro:conver}
In Actor-Critic algorithm, gradient ascent on $Q(s,a)\pi_{\theta_{i}}(a|s)$ pushes the next updated policy $\pi_{\theta_{i+1}}$ in the direction provided by the $Q$ value function. In the presence of false positive rewards, the gradient of $Q(s,a)\pi(a|s)$ can be expressed as follows:  
\begin{flalign}
\label{eq:coverate_redu}
&\nabla_\theta Q_\phi(s,a)\pi_\theta(a|s) = \textit{const} \cdot \mathbb E [G^{\pi \in \Pi_L}] \nabla_\theta \pi_\theta(a|s) \nonumber \\  &\qquad + ( \mathbb E [G^{\pi \in \Pi_G}] - \mathbb E [G^{\pi \in \Pi_L}]) P(\pi \in \Pi_G) \nabla_\theta \pi_\theta(a|s) \nonumber
\end{flalign}
where $\Pi_G$ is the set of acceptable policies; $\Pi_L$ is an arbitrary set of suboptimal policies that are partially consistent; $G^\pi$ is the cumulative rewards by executing policy $\pi$ in one episode.
\end{theorem}

\begin{table*}[t]
\centering
\caption{Reward rules for the simulated LRS model: Rule 1 rewards only fully matched trajectories, while Rule 2 allows partial matches with reduced reward magnitude; Rule 3 further relax the temporal ordering constraints and allows for rewards to be obtained out of order.}
\renewcommand{\arraystretch}{1}
\label{tab:simsetting}
\begin{tabular}{l l p{3cm}} 
\toprule
  \multicolumn{1}{c}{Name} & \multicolumn{1}{c}{Rewarding Set}                                   & \multicolumn{1}{c}{Magnitude}                       \\ 
\hline
Rule 1: Fully Matched & $\{\tau^{\pi_{inst}} | \pi_{inst} \in \Pi_{C_a} \cap \Pi_{C_s} \cap \Pi_{C_t}\}$ & $r = 1.0$                                                  \\ 
\hline
Rule 2: Partially Matched & $\{\tau^{\pi_{inst}} | \pi_{inst} \in \Pi_{C_a} \cup \Pi_{C_s} \cap \Pi_{C_t}\}$ & $r = 0.5$ if $\pi_{inst}$ is partially matched  \\ 
\cline{1-2}
Rule 3: Relaxed Ordering & $\{\tau^{\pi_{inst}} | \pi_{inst} \in \Pi_{C_a} \cup \Pi_{C_s}\}$ & $r = 1.0$ if $\pi_{inst}$ is fully matched                   \\
\bottomrule
\end{tabular}
\end{table*}

The proof is in Appendix~\ref{sec:proofofreductionofconvergence}. Since the goal of the learning agent is to maximise $P(\pi \in \Pi_G)$ (i.e., to converge to an acceptable policy), we can see that the second term provides the target direction with rate $( \mathbb E [G^{\pi \in \Pi_G}] - \mathbb E [G^{\pi \in \Pi_L}])$. Therefore, the ascent rate will decrease when the expectation of the rollout cumulative false positive rewards gets higher. Moreover, the first term $\textit{const} \cdot \mathbb E [G^{\pi_L}] \nabla_\theta \pi_\theta(a|s)$ can be regarded as the deviation of target direction. It shows that the level of deviation is also positively proportional to the magnitude of expectation of the rollout cumulative false positive rewards. Therefore, rewarding partially matched trajectories is shown to reduce the convergence rate of the RL algorithm. Note that in this setting, rewards for partially matched trajectories are of a lower magnitude (i.e., $\mathbb E [G^{\pi \in \Pi_G}] > \mathbb E [G^{\pi \in \Pi_L}]$). Consequently, LRS models can theoretically converge, despite the existence of false positive rewards. However, our empirical evaluation revealed that agents trained with LRS rewards exhibit slower convergence rates compared to pure RL agents.

\section{Experiments and Result Analysis}

\subsection{Experimental Setup}

We conduct experiments in the Atari game Montezuma's Revenge, a challenging benchmark with sparse rewards, to evaluate the impact of rewarding partially matched trajectories on LRS agents. We conduct our experiments in three distinct rooms (A1, A2, and B3) to test the consistency of our analysis across different tasks. We collected top-rated instruction sentences from game forums relating to these levels. Our RL policy model is based on PPO and RND RL algorithm \citep{burda2018exploration}. Specifically, RND algorithm gives auxiliary rewards when agents reach unseen states, which is a widely accepted curiosity-driven exploration strategy. We use Area Under the Learning Curve (AUC) metric. This AUC metric, having been previously endorsed in studies like \citet{goyal2019using}, emerges as a more fitting indicator of the convergence rate for RL models. For more details of AUC, refer to Appendix~\ref{sec:appendiximpledetail}. 

\subsubsection{Non-Simulated LRS Model}

While earlier models cited in Section~\ref{sec:relatedwork} were often tailored to domain-specific tasks and limited by the technologies of their time, they all operate within a similar LRS framework. Given this commonality, we felt it justified to construct our own model, enabling us to sidestep those domain and technological constraints. Our non-sim LRS model incorporates state-of-the-art sentence embedding models and visual encoders. Preliminary evaluations revealed that existing open-sourced pretrained multimodal models (e.g., BLIP \citep{li2023blip2} and mPLUG \citep{xu2023mplug2}) had a poor performance in encoding cartoon images. Addressing this, our LRS model implemented a trainable binary classification output layer, as depicted in Figure~\ref{fig:twotypeoflrs_b}. Specifically, the model uses a T5 sentence transformer \citep{raffel2020exploring} as the language encoder; a masked auto-encoding pre-training objective \citep{seo2022masked}; soft-discretisation with YOLOv7 \citep{wang2022yolov7} for the visual observation encoder; and GATED XATTN module \citep{alayrac2022flamingo} as the binary classifier backbone. The fine-tuning involved 5000 annotated Montezuma's Revenge clips from Amazon Mechanical Turk, supplemented by contrastive learning using hard negatives. Ablation studies and more details can be found in Appendix~\ref{sec:appendixadditionalstudies}. It is important to clarify that the issue of rewarding partially matched trajectories is not confined to particular choices of LRS encoders. Instead, it arose from scalar compression and smooth gradient-based learning, as discussed in Section~\ref{sec:brittleoflrs}. 

\subsubsection{Simulated LRS Model}
In our efforts to assess the impact of rewarding partially matched trajectories without the interference of other factors such as domain shift, poor data quality, and approximation error due to the choice of multimodal architectures, we devised a simulated LRS model. The model compares the sequence of past actions and states of the agent with predefined target trajectory sets that map to each instruction sentence. It subsequently assigns rewards in the following modes (see Table~\ref{tab:simsetting}):
\begin{itemize}
    \item Rule 1 dispenses rewards exclusively for chronologically arranged trajectories that align perfectly with each instruction sentence.
    \item Rule 2 allows for partial trajectory matches but with diminished reward magnitude, while preserving the temporal order.
    \item Rule 3 further removes the temporal ordering constraints and permits rewards even if obtained out of sequence.
\end{itemize}    

Our careful design ensures that the simulated LRS model already represents the upper bound, making it a valuable tool to demonstrate the issue caused by rewarding partially matched trajectories. Further implementation details are in Appendix~\ref{sec:appendiximpledetailforsimu}.


\subsection{Main Results}
\label{subsec:mainresult}

\begin{table*}[t]
\centering
\caption{Performance of agents in Montezuma's Revenge game, measured by AUC (higher is better), as well as success rate for perturbed rewards tolerance. Baseline denoted as B. $\star$ indicates statistical significance (p $<$ 0.05).}
\label{tab:mainresult}
\begin{tblr}{width = \linewidth,colspec = {Q[473]Q[346]Q[106]},row{1} = {p},hline{1-2,9} = {-}{},hline{4,6} = {-}{dashed}}
Model                    & AUC             & SR    \\
PPO \citep{schulman2017proximal} & all failed      & 0\%   \\
PPO+RND (B) \citep{burda2018exploration}             & 0.550$\pm$0.066 & 100\% \\
PPO + Sim LRS Rule 1     & 0.287$\pm$0.048 & 100\% \\
PPO+RND + Non-Sim LRS    & 0.069$\pm$0.139 $\star$& 16.7\%   \\
PPO+RND + Sim LRS Rule 1 & 0.608$\pm$0.073 & 100\% \\
PPO+RND + Sim LRS Rule 2 & 0.183$\pm$0.187 $\star$& 73.3\%  \\
PPO+RND + Sim LRS Rule 3 & 0.051$\pm$0.116 $\star$& 16.7\%   
\end{tblr}
\end{table*}

Results are reported in Table~\ref{tab:mainresult}. Several important observations can be made:

1. LRS techniques have shown enhanced performance compared to weak RL baselines like PPO. While agents trained solely with PPO struggled to navigate through all three rooms, incorporating LRS into PPO (i.e., PPO+LRS) did find solutions, albeit with modest success rates. However, the non-simulated LRS agent remained challenged by room A2, a task that necessitates intricate sequential decision-making.

\begin{figure*}[t]
    \centering
     \begin{subfigure}[b]{0.24\textwidth}
         \centering
         \includegraphics[width=\textwidth, height=3.8cm]{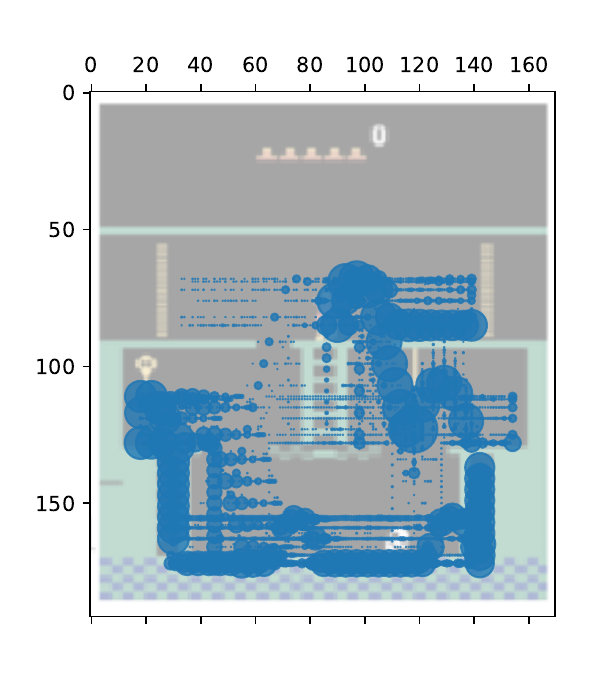}
         \caption{PPO+RND agent}
         \label{fig:heatmapa} 
     \end{subfigure}
     \hfill
     \begin{subfigure}[b]{0.24\textwidth}
         \centering
         \includegraphics[width=\textwidth, height=3.8cm]{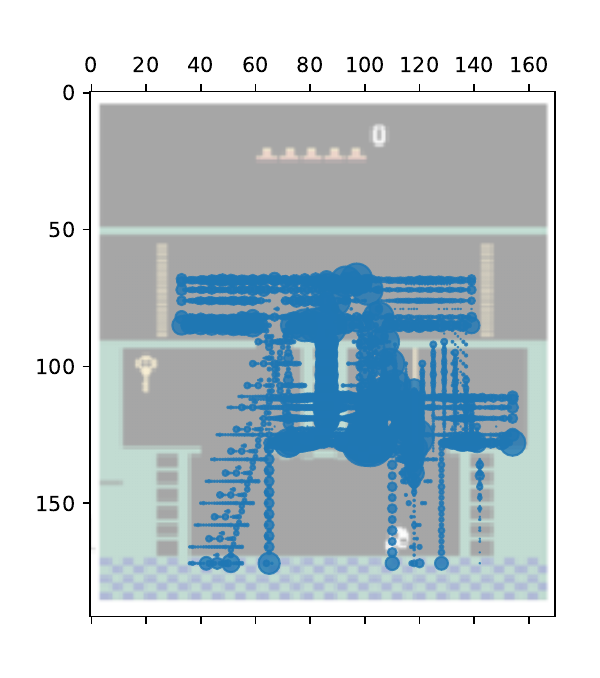}
         \caption{Non-Sim LRS agent}
         \label{fig:heatmapb} 
     \end{subfigure}
     \hfill
     \begin{subfigure}[b]{0.24\textwidth}
         \centering
         \includegraphics[width=\textwidth, height=3.8cm]{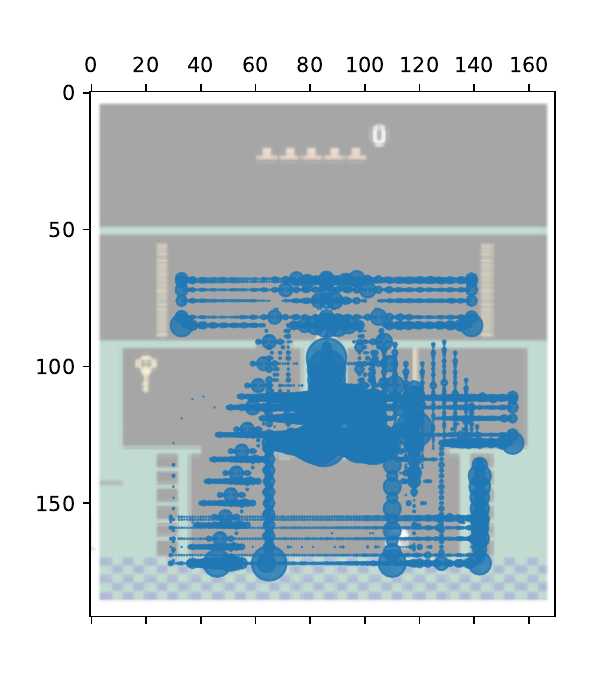}
         \caption{Sim LRS (Rule 2)}
         \label{fig:heatmapd} 
     \end{subfigure}
     \hfill
     \begin{subfigure}[b]{0.24\textwidth}
         \centering
         \includegraphics[width=\textwidth, height=3.8cm]{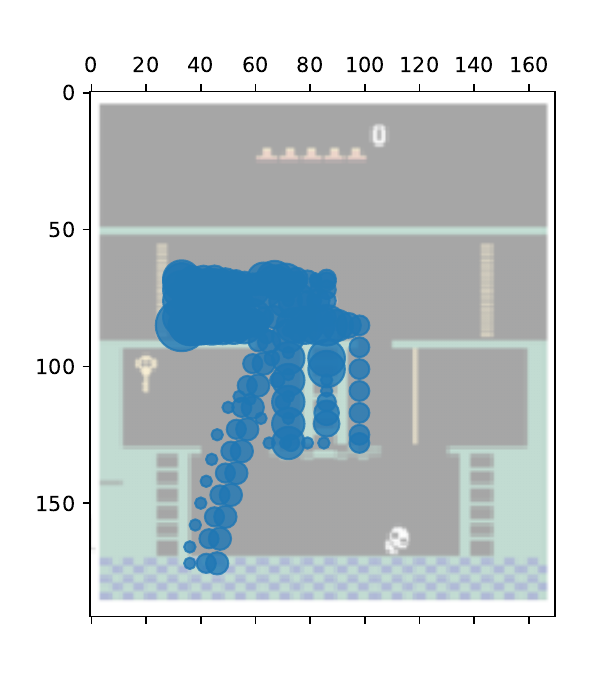}
         \caption{Sim LRS (Rule 3)}
        \label{fig:heatmapc}   
     \end{subfigure}

    \caption{Movement heatmap for agents in different settings}
    \label{fig:heatmap}
\end{figure*}

2. In our qualitative assessment of the Room A2 task, both non-simulated and simulated LRS agents displayed instances of receiving false positive rewards, as depicted in Figure~\ref{fig:lrsfalsepositiveexample}. These misleading rewards prompted the LRS agents to persistently adopt suboptimal strategies, hindering them from completing the desired long trajectories. Additionally, the movement heatmaps in Figures \ref{fig:heatmapb} and \ref{fig:heatmapd} showed a noticeable similarity, indicating that the simulated LRS model (Rule 2) accurately captured the behavior of the real LRS model. The use of the simulated LRS model thus provided a valuable means of highlighting practical issues with LRS implementations.

3. The AUC value of both non-simulated and simulated LRS agents (Rule 2 and Rule 3) underperformed versus pure PPO+RND agents. This deterioration of learning efficiency is statistically significant. Both non-simulated and simulated agents were unable to reach a goal state in certain training runs. In particular, the success rate of non-simulated LRS agents experienced a notable drop, indicating the destabilizing effect of false positive rewards on the learning process.


4. The algorithm completely failed to converge when LRS models neglected the temporal ordering (i.e., Rule 3). The heatmap in Figure~\ref{fig:heatmapc} showed that the learning agent kept pursuing rewards from the last instruction sentence (i.e., ``walk left to the door''). However, the agent did not escape the room by simply executing the last instruction because a key was required to unlock the door. It suggests that the agent tends to be trapped by local minima that are close to the initial state $s_0$ when temporal ordering is disregarded. 

\subsection{Analysis of Instructions of Varied Granularity}
\label{subsec:diffgran}
As part of the test, we also examined how different levels of granularity affect performance. We simulated a less detailed instruction in following two ways: 1) skipping intermediate steps and 2) neglecting an entire aspect of information, such as omitting actions details. The two types of less detailed instructions are illustrated in Table~\ref{tab:loweringdetail}. Communicating with a low granularity of information is very common as human communicators often believe that some intermediate steps are not important or they intentionally omit common-sense information under the assumption that they share the same context with their audience \citep{mitkov2014anaphora}. 

\begin{table*}
\centering
\caption{Less detailed instructions represented in trajectory sequences, with human readable examples.}
\label{tab:loweringdetail}
\begin{tblr}{width = \linewidth,colspec = {Q[30]Q[380]Q[527]},column{1} = {c},hlines,hline{1,4} = {-}{0.08em}}
Type & Representation                                          & Example                                                                                                                    \\
1    & $(s_0, a_n, s_n, ... , a_{kn}, s_{kn}), n,k\in \mathbb Z^+$ & Climb down the ladder \st{and walk right on the conveyor belt}, after that, jump to the yellow rope.                   \\
2    & $(s_0, s_1, s_2, ... , s_k), k\in \mathbb Z^+$          & \st{Climb down} To the ladder and \st{walk right on} to the conveyor belt, after that, \st{jump} to the yellow rope.
\end{tblr}
\end{table*}

\begin{figure}[t]
    \centering
    \includegraphics[width=0.8\columnwidth]{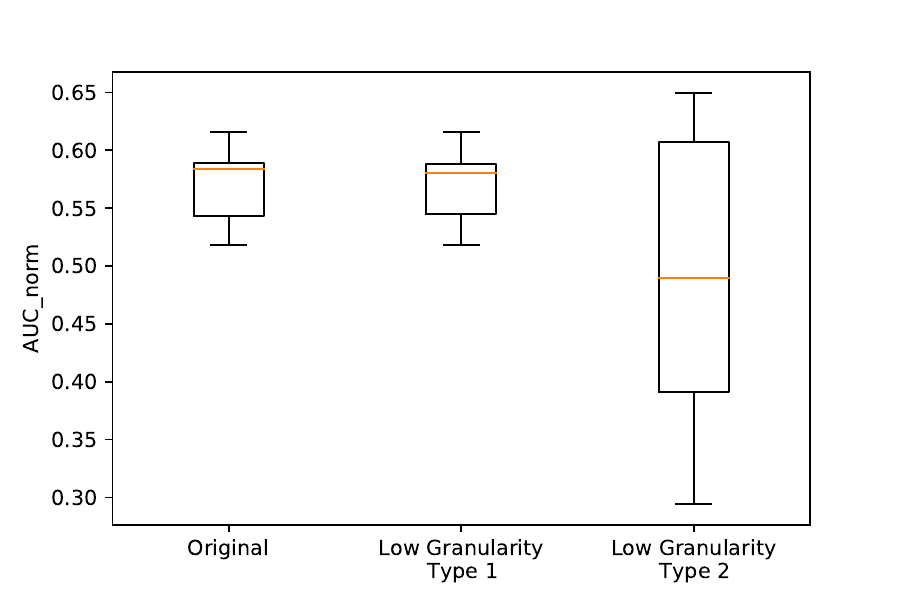}
    \caption{Comparison of the learning efficiency across different granularity -- Type 1 skips intermediate steps; Type 2 neglects a whole dimension.}
    \label{fig:loweringdetail}
\end{figure}

Our result showed that the two types of less detailed instructions led to different performance outcomes. As shown in Figure~\ref{fig:loweringdetail}, we observed no statistically significant deterioration in learning efficiency when intermediate steps are skipped. This suggested that RL algorithms are capable of interpolating missing information based on the given context and therefore can quickly recover from the missing intermediate steps. However, when removing a whole dimension of constraints, finding the goal would take more training time with a larger variance. This indicated that in such instances, the model essentially relearns from scratch and thereby lacking any prior commonsense knowledge to extrapolate the missing dimension of the information.

\subsection{Effectiveness of Language Reward Shaping}

Our results also revealed that agents with an ideal language reward shaping model might still learn more slowly than a vanilla PPO+RND RL agent in difficult environments. We attribute this observation to the fact that LRS agents are constrained to follow instructions that may not cover all the acceptable policies (i.e., $\Pi_{\mathit{INST}} \subset \Pi_G$). Therefore, the trade-off between a smoother value function manifold and the reduced size of the goal policy set can result in slower learning, as a large number of acceptable policies are pruned out. For instance, the heatmap in Figure~\ref{fig:heatmapa} showed that PPO+RND agent found an alternative solution path (i.e., directly jumping from the cliff) rather than the one suggested by the instruction (i.e., climbing down the ladder). Nevertheless, we speculate that LRS can have a positive effect when $|\Pi_{\mathit{INST}}| \approx |\Pi_G|$. This would require building an LRS model that can interpret more general language information, such as manuals and strategies, to cover almost all possible circumstances and solutions. This suggests the need for a more intelligent language model that can ground general information to agent actions.

\section{Conclusion}
\label{sec:conclusion}

The combination of a powerful foundation language backbone, such as T5 \citep{raffel2020exploring}, with state-of-the-art RL algorithms might be seen as a solution to effectively tackle sequential decision-making problems. However, our study highlights the need for a careful design of the knowledge transfer mechanism from large pretrained models to domain-specific models to achieve effective results. Specifically, we observed that 1) the challenges of LRS models when compressing high-dimensional vectors into scalars are not contingent upon specific language models or the quality of the training dataset; 2) LRS models that reward partially matched trajectories are statistically slower than pure RL agents in the Montezuma sparse reward environment. Furthermore, the introduction of reward perturbation by loosening task constraints presents a novel extension to the existing types of poisoned rewards and also highlights the unique challenges involved in using LRS. We stress that our goal is not to dismiss the concept of using language instructions to provide auxiliary rewards for training agents. Instead, we aim to shed light on the severe bottleneck faced by existing LRS approaches and the need for addressing these fundamental limitations. By recognizing and addressing these challenges, we can pave the way for more effective and robust LRS architectures in the future.

While we believe that LRS models which perform poorly in Montezuma would face even more significant convergence issues in more complex real-world scenarios, we acknowledge the limited scope of this work -- the impact of rewarding partially matched trajectories may vary depending on the tasks' difficulty levels. For instance, in the case of Montezuma room A1, which is a simpler task compared to room A2 and B3, we observed that LRS agents learned faster compared to PPO+RND agents. Nevertheless, the results from the simulated LRS models indicated that the convergence was still severely hampered by the presence of false positive rewards. Specifically, we observed that the influence of false positive rewards may be alleviated in environments with smaller action spaces, as demonstrated by \citet{du2023guiding}. Moreover, we recognise that the benefits of suboptimal LRS can outweigh its drawbacks in scenarios where exploration is undesirable. For instance, 1) in dangerous real-world environments, exploration can be costly; or 2) in tasks that require repeated actions, naive exploration may not provide additional benefits.

We suggest two directions for improvement. First is to consider the compositionality of the language instructions. One potential approach is to turn the scalar reward signals into vectorised multi-label rewards. This could mitigate ambiguity in reward signals by explicitly representing the level of satisfaction for different types of constraints. Second, the type of rewarding shall not be limited to immediate numerical rewards for RL algorithms. We could formulate the learning problem as a trajectory optimisation problem (e.g., \citet{janner2022planning}), using language instructions to directly upvote a set of matched trajectories. This may help to avoid cumulative rollout errors caused by single-step reward models, and thereby improving the learning efficiency.

\bibliography{aaai24}

\appendix
\onecolumn
\section{Implementation Details}
\label{sec:appendiximpledetail}
\textbf{Computation} The experiments were tested on a single NVIDIA GeForce RTX 3090 GPU, with 8 CPU cores. Each training trial is run for a total of 20 million frames. Average duration for each trial is around 1 hour. 

\noindent{\textbf{Environment hyperparameter}} 

\begin{table}[H]
\centering
\caption{Hyperparameter of the Montezuma's Revenge game environment}
\label{tab:envhyper}
\begin{tabular}{c|c}
Hyperparameter                & Value                                            \\ 
\hline
Grey-scaling                  & True                                             \\
Observation Resize            & (84,84)                                          \\
Language reward clipping      & {[}0,1]                                          \\
Intrinsic reward clipping     & {[}0,5]                                          \\
Environmental reward clipping & {[}0,1]                                          \\
Max frames per episode        & 1200                                             \\
Terminal on loss of life      & False                                            \\
Max and skip frames           & 4                                                \\
Random starts                 & 30                                               \\
Sticky action probability     & 0.25                                             \\
Frames stacked                & 4                                                \\
Size of action space          & 8                                                \\
image normalization mean      & {[}0.485 * 255.0, 0.456 * 255.0, 0.406 * 255.0]  \\
image normalization std       & {[}0.229 * 255.0, 0.224 * 255.0, 0.225 * 255.0]  \\
random seed                   & \{1,2,3,4,5,6,7,8,9,10\}                        
\end{tabular}
\end{table}

\noindent{\textbf{Model hyperparameter}} 

\begin{table}[H]
\centering
\caption{Hyperparameter of the PPO+RND policy model}
\label{tab:modelhyper}
\begin{tabular}{c|c}
Hyperparameter                  & Value   \\ 
\hline
learning rate                   & 1.0e-4  \\
Rollout length                  & 128     \\
env gamma                       & 0.99    \\
language gamma                  & 0.99    \\
intrinsic gamma                 & 0.99    \\
use GAE                         & True    \\
GAE lambda                      & 0.95    \\
mini batch                      & 4       \\
Entropy coefficient             & 0.001   \\
env and lang reward coefficient & 3.0     \\
intrinsic reward coefficient    & 1.0     \\
Size of action space            & 8       \\
Policy architecture             & CNN     \\
batch size                      & 256     \\
number of parallel environments & 8      
\end{tabular}
\end{table}

\noindent{\textbf{The task environment Room A2 and the associated natural language instructions}} 
\begin{figure}[H]
    \includegraphics[width=0.80\textwidth]{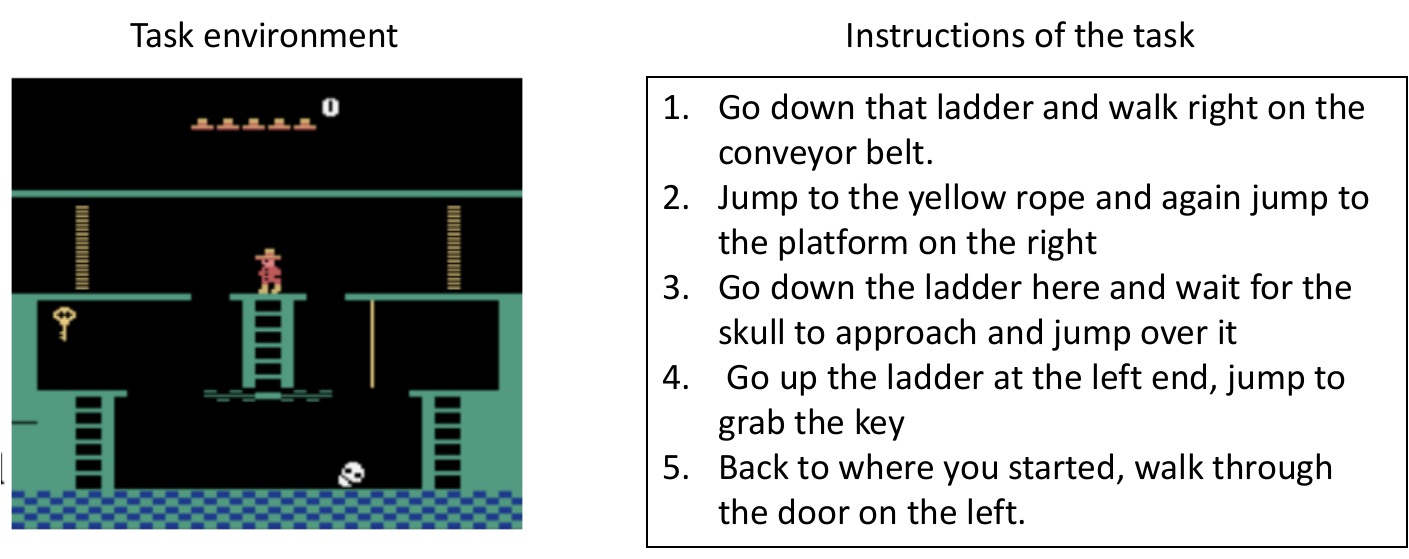}
    \centering
\end{figure}

\noindent{\textbf{The task environment Room A1 and the associated natural language instructions}} 
\begin{figure}[H]
    \includegraphics[width=0.80\textwidth]{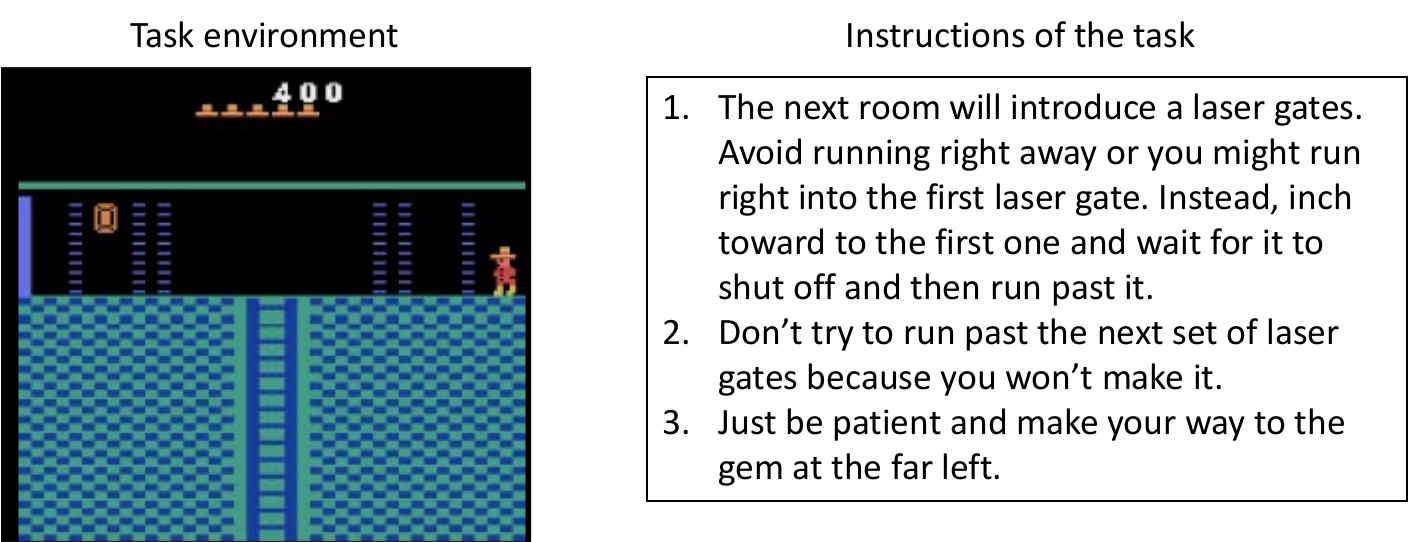}
    \centering
\end{figure}

\noindent{\textbf{The task environment Room B3 and the associated natural language instructions}} 
\begin{figure}[H]
    \includegraphics[width=0.80\textwidth]{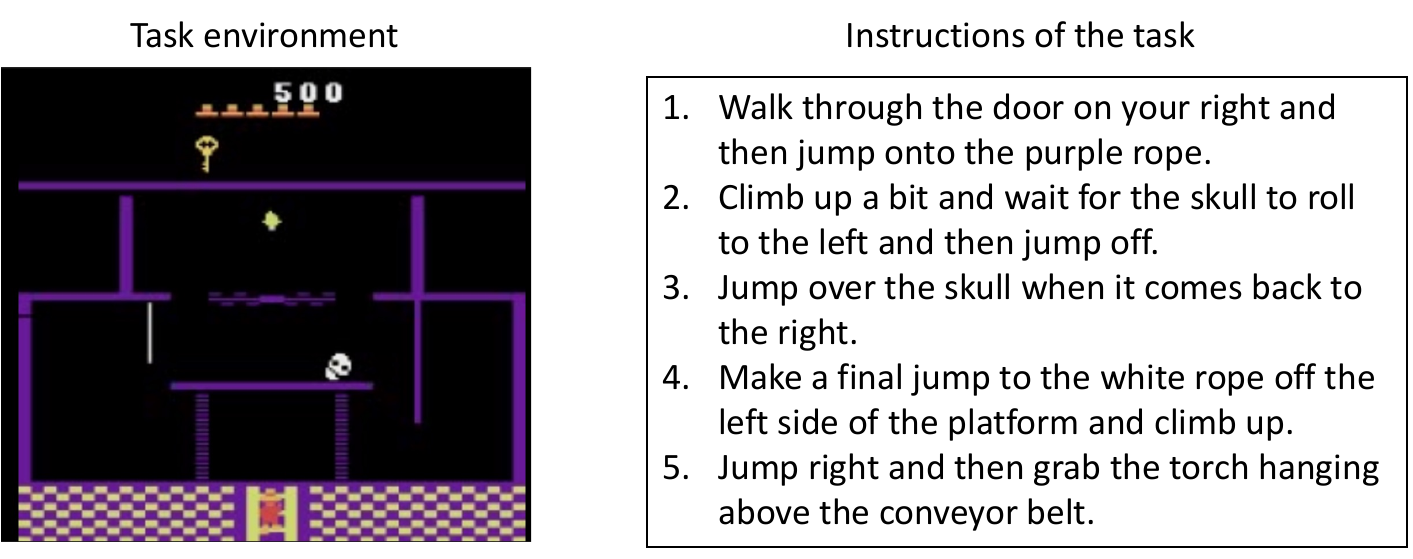}
    \centering
\end{figure}

\textbf{Evaluation Metrics} We set a cap on the number of wins an agent can achieve during training at 1500. This means that once the agent reaches this limit, the training process will automatically stop. We assume that the training becomes stable after reaching this limit. The evaluation metric is the area under the curve (AUC) normalised by the length of training episodes. Specifically, we calculate the AUC by summing up the number of wins achieved at each time step $i$ during training and dividing it by the total number of training episodes $T$. It can be expressed as:

\begin{flalign}
    \text{AUC} = \frac{\sum_{i=0}^{T} \text{no. of win}_i}{T} \nonumber
\end{flalign}

\section{Implementation details for Simulated LRS models}
\label{sec:appendiximpledetailforsimu}

For each environment tested, we first defined sets of target trajectories $\tau_i^{\Pi_{G}}$ that correspond to each instruction sentence $D_i$. During policy training, the simulated LRS model receives the current trajectory of the agent $\tau^\pi = \{s_0, a_0, s_1, ..., a_{t-1}, s_t\}$ at each time $t$. It then compares $\tau^\pi$ with the sets of predefined target trajectories $\{\tau_i^{\Pi_{G}}\}$ and provides immediate rewards according to reward rules stated in Table~\ref{tab:simsetting}. 

Rule 1 simulates a type of LRS model that does not reward partially matched trajectories. Rule 2 and Rule 3 simulates LRS models that reward partially matched trajectories. The difference between the two is that in Rule 2, the temporal ordering constraints are strictly maintained by heuristics, while in Rule 3, partially matched trajectories that do not satisfy temporal constraints can also be rewarded. To simplify our analysis, we focus solely on one type of temporal constraint, which is the order in which each instruction sentence is executed. Therefore, temporal constraints are met if the agent's trajectory matches the correct order of the target trajectory sequence. For example, if $\tau^\pi$ matches with $\tau_i^{\Pi_{G}} ++ \tau_{i+1}^{\Pi_{G}}$, we say $\tau^\pi$ fulfils the temporal constraints. 

It is important to note that the reward rule in Table~\ref{tab:simsetting} ensures that the rollout cumulative rewards of partially matched trajectories will always be smaller than that of fully matched ones. This ensures that the RL algorithm will converge eventually in theory. 

\section{Detailed Experiment Result Tables}
\label{sec:appendixaddexpresultstab}

In this study, we focus on the PPO+RND RL backbone, given that LRS methods are typically harnessed for tasks with sparse reward signals, a scenario where the PPO algorithm alone has been proven ineffective \citep{schulman2017proximal, burda2018exploration}. The vanilla ``PPO+RND'' model serves as the baseline, and we compare it with ones supplemented by various LRS models. The inclusion of ``PPO + Sim LRS Rule 1'' in Table~\ref{tab:mainresulta2} is merely to illustrate that under ideal conditions, LRS is intended to support PPO in overcoming its ineffectiveness with sparse reward tasks. By showing that the PPO+RND model can be significantly hindered by issue of rewarding partially matched trajectories, we demonstrate the brittleness of the existing LRS techniques. The PPO algorithm will only magnify this issue, given its comparative ineffectiveness against RND when faced with sparse-reward settings.

\begin{table}[H]
\centering
\caption{Learning efficiency of agents in Room A2 of Montezuma's Revenge game, measured by AUC (higher is better), and success rate (for perturbed rewards tolerance). Baseline denoted as B. $\star$ indicates statistical significance (p $<$ 0.05).}
\label{tab:mainresulta2}
\begin{tblr}{
  width = \linewidth,
  colspec = {Q[319]Q[244]Q[228]Q[146]},
  row{1} = {c},
  cell{2}{2} = {c},
  cell{2}{3} = {c},
  cell{2}{4} = {c},
  cell{3}{2} = {c},
  cell{3}{3} = {c},
  cell{3}{4} = {c},
  cell{4}{2} = {c},
  cell{4}{3} = {c},
  cell{4}{4} = {c},
  cell{5}{2} = {c},
  cell{5}{3} = {c},
  cell{5}{4} = {c},
  cell{6}{2} = {c},
  cell{6}{3} = {c},
  cell{6}{4} = {c},
  cell{7}{2} = {c},
  cell{7}{3} = {c},
  cell{7}{4} = {c},
  cell{8}{2} = {c},
  cell{8}{3} = {c},
  cell{8}{4} = {c},
  hline{1,9} = {-}{0.08em},
  hline{2} = {-}{},
  hline{4,6} = {-}{dashed},
}
Model                    & AUC (higher is better) & p-value w.r.t. B & success rate \\
PPO \citep{schulman2017proximal}                     & all failed             & -                & 0\%          \\
PPO+RND (B)              & 0.633$\pm$0.038        & -                & 100\%        \\
PPO + Sim LRS Rule 1     & 0.287$\pm$0.048        & 4.5e-05$\star$                & 100\%        \\
PPO+RND + Non-Sim LRS    & all failed             & $\star$                & 0\%          \\
PPO+RND + Sim LRS Rule 1 & 0.571$\pm$0.028        & 8.5e-04$\star$   & 100\%        \\
PPO+RND + Sim LRS Rule 2 & 0.165$\pm$0.206        & 4.3e-06$\star$   & 70\%         \\
PPO+RND + Sim LRS Rule 3 & all failed             & $\star$                & 0\%          
\end{tblr}
\end{table}

\begin{table}[H]
\centering
\caption{Learning efficiency of agents in Room A1 of Montezuma's Revenge game, measured by AUC (higher is better), and success rate (for perturbed rewards tolerance). Baseline denoted as B. $\star$ indicates statistical significance (p $<$ 0.05).}
\label{tab:mainresulta1}
\begin{tblr}{
  width = \linewidth,
  colspec = {Q[344]Q[259]Q[179]Q[155]},
  row{1} = {c},
  cell{2}{2} = {c},
  cell{2}{3} = {c},
  cell{2}{4} = {c},
  cell{3}{2} = {c},
  cell{3}{3} = {c},
  cell{3}{4} = {c},
  cell{4}{2} = {c},
  cell{4}{3} = {c},
  cell{4}{4} = {c},
  cell{5}{2} = {c},
  cell{5}{3} = {c},
  cell{5}{4} = {c},
  cell{6}{2} = {c},
  cell{6}{3} = {c},
  cell{6}{4} = {c},
  cell{7}{2} = {c},
  cell{7}{3} = {c},
  cell{7}{4} = {c},
  hline{1,8} = {-}{0.08em},
  hline{2} = {-}{},
  hline{4-5} = {-}{dashed},
}
Model                    & AUC (higher is better) & p-value w.r.t. B & success rate \\
PPO                      & all failed             & -                & 0\%          \\
PPO+RND (B)              & 0.201$\pm$0.038        & -                & 100\%        \\
PPO+RND + Non-Sim LRS    & 0.174$\pm$0.220        & 0.32                & 40\%          \\
PPO+RND + Sim LRS Rule 1 & 0.433$\pm$0.085        & 0.99             & 100\%        \\
PPO+RND + Sim LRS Rule 2 & 0.225$\pm$0.229        & 0.61             & 50\%         \\
PPO+RND + Sim LRS Rule 3 & 0.152$\pm$0.201        & 0.14             & 50\%         
\end{tblr}
\end{table}

\begin{table}[H]
\centering
\caption{Learning efficiency of agents in Room B3 of Montezuma's Revenge game, measured by AUC (higher is better), and success rate (for perturbed rewards tolerance). Baseline denoted as B. $\star$ indicates statistical significance (p $<$ 0.05).}
\label{tab:mainresultb3}
\begin{tblr}{
  width = \linewidth,
  colspec = {Q[319]Q[244]Q[228]Q[146]},
  row{1} = {c},
  cell{2}{2} = {c},
  cell{2}{3} = {c},
  cell{2}{4} = {c},
  cell{3}{2} = {c},
  cell{3}{3} = {c},
  cell{3}{4} = {c},
  cell{4}{2} = {c},
  cell{4}{3} = {c},
  cell{4}{4} = {c},
  cell{5}{2} = {c},
  cell{5}{3} = {c},
  cell{5}{4} = {c},
  cell{6}{2} = {c},
  cell{6}{3} = {c},
  cell{6}{4} = {c},
  cell{7}{2} = {c},
  cell{7}{3} = {c},
  cell{7}{4} = {c},
  hline{1,8} = {-}{0.08em},
  hline{2} = {-}{},
  hline{4-5} = {-}{dashed},
}
Model                    & AUC (higher is better) & p-value w.r.t. B & success rate \\
PPO                      & all failed             & -                & 0\%          \\
PPO+RND (B)              & 0.817$\pm$0.101        & -                & 100\%        \\
PPO+RND + Non-Sim LRS    & 0.033$\pm$0.099       & 1.1e-12$\star$   & 10\%         \\
PPO+RND + Sim LRS Rule 1 & 0.821$\pm$0.090        & 0.53             & 100\%        \\
PPO+RND + Sim LRS Rule 2 & 0.161$\pm$0.103        & 3.5e-11$\star$   & 100\%        \\
PPO+RND + Sim LRS Rule 3 & all failed             & $\star$                & 0\%          
\end{tblr}
\end{table}

\section{Proofs}
\label{sec:appendixproofs}

\subsection{Proof of convergence guarantee of the potential-based perfect LRS (Proposition~\ref{prop:conver})}
The proof is derived firmly anchored in the foundational work present in \citet{wiewiora2003principled} and \citet{okudo2021subgoal}. The convergence guarantee property is only applicable to on-policy RL algorithms such as SARSA and PPO.
We start by proving $F(s_t,a_t, s_{t-1}, a_{t-1}) = \Phi(s_t,a_t) -\gamma^{-1} \Phi(s_{t-1}, a_{t-1})$ is a potential-based reward shaping function (i.e., the optimal policy in the old setting remains optimal when the reward function is modified by adding the reward shaping function). We first of all write down the optimal Q-value of the original MDP:
\begin{equation}
    Q^*(s,a) = \mathbb E[\sum_{t=0}^\infty \gamma^t R^{\text{env}}_t | s_0 =s, \pi = \pi^*]
\end{equation}
We now write down the optimal Q-value when the reward function is modified with the shaping reward, we denote it as $Q_\Phi^*$:
\begin{equation}
    \begin{aligned}[b]
        Q_\Phi^*(s,a) &= \mathbb E[\sum_{t=0}^\infty \gamma^t (R^{\text{env}}_t + F_t)]\\
        & = \mathbb E [\sum_{t=0}^{\infty} \gamma^t(R^{\text{env}}_t + \Phi(s_t,a_t) - \gamma^{-1} \Phi(s_{t-1}, a_{t-1}))] \\
        & = \mathbb E[\sum_{t=0}^\infty \gamma^t R^{\text{env}}_t]  + \mathbb E[\sum_{t=0}^\infty \gamma^t \Phi(s_t,a_t)] - \mathbb E[\sum_{t=-1}^\infty \gamma^t \Phi(s_t,a_t)] \\
        & = \mathbb E[\sum_{t=0}^\infty \gamma^t R^{\text{env}}_t] - \gamma^{-1} \mathbb E [\Phi(s_{-1}, a_{-1})] \\
        & = Q^*(s,a)- \gamma^{-1} \mathbb E [\Phi(s_{-1}, a_{-1})]
    \end{aligned}
\end{equation}
The term $ \mathbb E [\Phi(s_{-1}, a_{-1})]$ refers to the expected  value of the potential function $\Phi$ of the previous state and action, given $\pi$. They are considered as the exploration history and will not impact the current update when RL algorithms are on-policy. Thus the optimal policy remains. Since instruction guides the learning agent to reach the final goal, it can be seen as telling the agent what subgoals are required in order to reach the final goal. the potential function can be set as $\Phi(s_t, a_t) = \alpha \times c(s_t, a_t)$, where $\alpha$ is a hyperparameter and $c(s_t, a_t)$ is a function that outputs the sequence number of the subgoal that the $(s_t, a_t)$ tuple contributes to. Therefore, the final shaping reward function $F$ can be seen as measuring the increment of the achieved subgoals from the last state. Thus, we proved that LRS function is approximately a potential-based shaping function and the optimal policy invariance is guaranteed. 

\textbf{Clarification:} It's crucial to recognize that while the ``look-back'' potential-based shaping function may appear non-Markovian -- given its reliance on preceding states and actions -- its true essence lies in a specific constraint: the shaping reward is tailored exclusively for on-policy RL algorithms such as PPO and RND. This distinction ensures that the term $\mathbb E [\Phi(s_{-1}, a_{-1})]$ is interpreted solely as a record of exploration history, without influencing the contemporaneous update of the on-policy model.

The reason why the ``look-ahead'' reward function $F(s,a,s',a') = \gamma\Phi(s',a') - \Phi(s,a)$ is not applicable here is because it imposes a strong assumption requiring the potential function to remain deterministic and stable in order to have the optimal policy unchanged. However, this condition is not satisfied for LRS because the reward function undergoes updates in each iteration. \citet{wiewiora2003principled} provides a comprehensive explanation regarding this matter.

\subsection{Proof of the Reduction of Convergence Rate}
\label{sec:proofofreductionofconvergence}

Specifically, the update rule of Actor-Critic algorithm is: 

\begin{itemize}
    \item \textbf{Critic}: 
    \begin{equation}
        \phi \leftarrow \phi - \alpha_\phi \nabla_\phi (\delta)^2
    \end{equation}
    where 
    \begin{description}
        \item $\delta = \mathbb E_{\pi_{\theta}}[G^{\pi_{\theta}} - Q_\phi (s,a)]$ is the Monte Carlo (MC) estimation error,
        \item $G^{\pi_{\theta}}$ is the rollout cumulative rewards from the trajectory $\tau^{\pi_{\theta}}$ generated from $\pi_{\theta}$.
    \end{description}
    \item \textbf{Actor}: 
    \begin{equation}
        \theta \leftarrow \theta + \alpha_\theta \frac{Q_\phi(s,a) \nabla_\theta \pi_\theta(a|s)} {\pi_\theta(a|s)}
    \end{equation}
\end{itemize}

We need to make the following assumptions to simplify the theoretical analysis: 

\begin{enumerate}
    \item We use $\Pi_G$ to represent the set of acceptable policies and use $\Pi_L$ as an arbitrary set of suboptimal policies that are partially consistent with the instruction. For simplicity, we also assume $\Pi_{L} \cap \Pi_{G} = \varnothing$, meaning that $\Pi_L$ cannot reach the goal state.

    \item Assume the policy class parameterized by $\theta$ should be expressive enough to capture optimal or near-optimal policies, and the policy is initialised randomly from uniform distribution, i.e., $\pi_{\theta_{0}} \sim \mathcal U(\Pi)$. Meanwhile, $Q_{\phi_{0}}(s, a) = 0, \forall_{s\in S} \forall_{a\in A}$, where $Q$-value function measured the expected discounted cumulative reward given the state $s$ and action $a$. 

    \item We assume that $|\Pi_{G}|$ and $|\Pi_L|$ is a fixed number predefined by the task environment while $\mathbb E[G^\pi| \pi \in \Pi_{L}]$ is non-zero as partially matched trajectories are rewarded.
\end{enumerate}

Since the update rule for $Q$-value is a gradient descent on $\| \mathbb E [G^{\pi_\theta} - Q_\phi(s,a)] \|^2$ and also we have that $\{\Pi_{L}, \Pi_{G}, \Pi \setminus (\Pi_{L} \cup \Pi_{G})\}$ is a countable partition of the policy universe $\Pi$, the updated $Q$-value will approach as follows:
\begin{flalign}
   &&             & \quad Q_\phi(s,a) \rightarrow \mathbb E [G^{\pi_\theta}] && \nonumber \\ 
   &&             &=          P(\pi_\theta \in \Pi_{L}) \cdot \mathbb E [G^{\pi_\theta} | \pi_\theta \in \Pi_{L}] +  P(\pi_\theta \in \Pi_{G}) \cdot \mathbb E [G^{\pi_\theta} | \pi_\theta \in \Pi_{G}] \nonumber \\
   &&             &\qquad + P(\pi_\theta \in \Pi \setminus (\Pi_{L} \cup \Pi_{G}) ) \cdot \mathbb E [G^{\pi_\theta} | \pi_\theta \in \Pi \setminus (\Pi_{L} \cup \Pi_{G})]     && \nonumber \\
   &&             &=          P(\pi_\theta \in \Pi_{L}) \cdot \mathbb E [G^{\pi_\theta} | \pi_\theta \in \Pi_{L}]  +  P(\pi_\theta \in \Pi_{G}) \cdot \mathbb E [G^{\pi_\theta} | \pi_\theta \in \Pi_{G}]
\end{flalign}
Given that the update rule for policy $\pi$ is the gradient ascent on $Q_\phi(s,a)\pi_\theta(a|s)$, we have the following:
\begin{flalign}
    &&              &\quad     \nabla_\theta Q_\phi(s,a)\pi_\theta(a|s)          &&  \nonumber \\
    &&                                          &= (\nabla_\theta P(\pi_{\theta_{old}} \in \Pi_{L}) \cdot \mathbb E [G^{\pi_{\theta_{old}}} | \pi_{\theta_{old}} \in \Pi_{L}] \pi_\theta(a|s)) && \nonumber \\
    &&                                          &\qquad + (\nabla_\theta P(\pi_{\theta_{old}} \in \Pi_{G}) \cdot \mathbb E [G^{\pi_{\theta_{old}}} | \pi_{\theta_{old}} \in \Pi_{G}] \pi_\theta(a|s)) && \nonumber \\
    &&                                          &= P(\pi_{L}) \cdot \mathbb E [G^{\pi_{L}}] \nabla_\theta \pi_\theta(a|s) + P(\pi_{G}) \cdot \mathbb E [G^{\pi_{G}}] \nabla_\theta \pi_\theta(a|s) &&  \nonumber \\
    &&                                          &= (1 - P(\pi_{G}) - P(\pi_\theta \in \Pi \setminus (\Pi_{L} \cup \Pi_{G}) )) \cdot \mathbb E [G^{\pi_{L}}] \nabla_\theta \pi_\theta(a|s) + P(\pi_{G}) \cdot \mathbb E [G^{\pi_{G}}] \nabla_\theta \pi_\theta(a|s) && \nonumber \\
    &&                                          &= \textit{const} \cdot \mathbb E [G^{\pi_L}] \nabla_\theta \pi_\theta(a|s) + ( \mathbb E [G^{\pi_G}] - \mathbb E [G^{\pi_L}]) P(\pi_G) \nabla_\theta \pi_\theta(a|s)
\end{flalign}   

\textbf{Justification of the assumptions:}
\begin{itemize}
    \item Regarding the non-zero probability of recovering the optimal policy at initialization, it is standard in theoretical analyses to assume a uniform distribution of a random variable at initialization (see \citep{agarwal2021theory}). This assumption does not contradict the conclusion about the convergence rate deterioration in Theorem~\ref{coro:conver}.
    \item In reference to the realizability condition implied by Assumption 2, the expressiveness of the policy class parameterized by $\theta$ is an underlying assumption for deep learning models, supported by the The Universal Approximation Theorem \citep{hornik1989multilayer}. 
\end{itemize}

\section{Temporal constraints from language instructions and Linear Temporal Logic}
\label{sec:appendixLTL}
As a start, we give the definition of atomic sentence -- in logic, an atomic sentence is a declarative sentence cannot be further broken down into other simpler sentences. For example, ``go down that ladder'' is an atomic sentence; ``walk right on the conveyor belt'' is also an atomic sentence, whereas ``go down that ladder and walk right on the conveyor belt'' is a molecular sentence in natural language. However, an atomic sentence in natural language usually does not correspond to a single $(s,a,s')$ state-action-next-state transition tuple. For instance, the atomic sentence ``go down that ladder'' may corresponds to a longer state-action sequence $(s_{\text{AtLadderTop}}, a_{\text{down}}, s_{\text{AtLadderMiddle}} ,a_{\text{down}}, s_{\text{AtLadderBottom}})$ because the ladder is long enough that it requires an agent to perform two steps of the ``down'' actions. Hence, for simplicity, we call the corresponding state-action sequence of an atomic sentence ``an atomic event'', labeled as $e$. Therefore, an agent trajectory $\tau$ can be written as a sequence of atomic events, i.e., $\tau = (e_0, e_1, e_2, ...)$, where $e_t$ is the atomic event at $t^{th}$ order.

The example of modular instruction sentence ``go down that ladder and walk right on the conveyor belt'' reflects a preference towards a set of trajectory that can be expressed as $(e_{a},..., e_{b})$, where $e_{a}$ is the event of going down the ladder and $e_{b}$ is the event of walking right on the conveyor belt, and there can be arbitrary number of unmentioned atomic events in between. However, if the sentence becomes ``go down that ladder after walking right on the conveyor belt'', the preferred trajectories becomes $(e_{b},..., e_{a})$. 

The replacement from the word ``and'' to the word ``after'' leads to change of the trajectory preference. In natural language, such a word indicates the temporal relation between the two mentioned atomic sentences. In order to avoid ambiguity, Linear Temporal Logic (LTL), a formal language, will be used to translate the natural language instructions because it can provide a mathematically precise notation for expressing the temporal relations between those atomic sentences. 

Specifically, LTL is the extension of propositional logic with two extra temporal modal operators: $\nextltl$ (next) and $\until$ (until). Given a set of the propositional atomic sentence symbols $P$, the syntax of an LTL formula is defined according to the following grammar: 
\begin{equation}
    \phi ::= \text{true} \quad \Big{|} \quad p \quad \Big{|} \quad \phi_1 \land \phi_2 \quad \Big{|} \quad \neg \phi \quad \Big{|} \quad \nextltl \phi \quad \Big{|} \quad \phi_1 \until \phi_2  \qquad ,\text{where $p\in P$}
\end{equation}
The $\until$ operator can further derive two more temporal modal operators $\eventually$ (eventually, i.e., sometimes in the future) and $\always$ (always, i.e., from now on forever):
\begin{equation}
    \begin{aligned}
        \eventually \phi &:= \text{true} \until \phi \\
        \always \phi &:= \neg \eventually \neg \phi 
    \end{aligned}
\end{equation}

An LTL formula can be satisfied by an infinite trajectory sequence of truth assignment of atomic event variables, i.e., $\tau = (e_0, e_1, e_2,...)$ for $E$, where $p\in e_i$ iff the propositional atomic sentence $p\in P$ holds at time step $i$. Formally, the satisfaction relation between a trajectory $\tau$ at time $i$ and an LTL formula $\phi$, denoted by $\langle \tau, i \rangle \models \phi$, is defined as follows:
\begin{equation}
    \begin{aligned}
        \langle \tau, i \rangle &\models true \\
        \langle \tau, i \rangle &\models p  &\text{iff}  \qquad & p\in e_i \quad \text{(i.e.,$e_i \models p$)}\\
        \langle \tau , i \rangle &\models \phi_1 \land \phi_2 &\text{iff}  \qquad & \langle \tau , i \rangle \models \phi_1 \quad \text{and} \quad \langle \tau , i \rangle\models \phi_2 \\
        \langle \tau , i \rangle &\models \neg \phi &\text{iff}  \qquad & \langle \tau , i \rangle \not\models \phi \\
        \langle \tau , i \rangle &\models \nextltl \phi &\text{iff}  \qquad & \langle \tau , i+1 \rangle \models \phi \\
        \langle \tau , i \rangle &\models \phi_1 \until \phi_2 &\text{iff}  \qquad & \exists j, i \leq j \quad \text{and} \quad \langle \tau , j \rangle \models \phi_2, \quad \text{and} \quad \langle \tau , k \rangle \models \phi_1 \quad \text{for all $k \in [i,j)$}\\
    \end{aligned}
\end{equation}

We give a sketch for the semantics of temporal modalities to better understand it.
\begin{figure}[H]
    \includegraphics[width=0.70\textwidth]{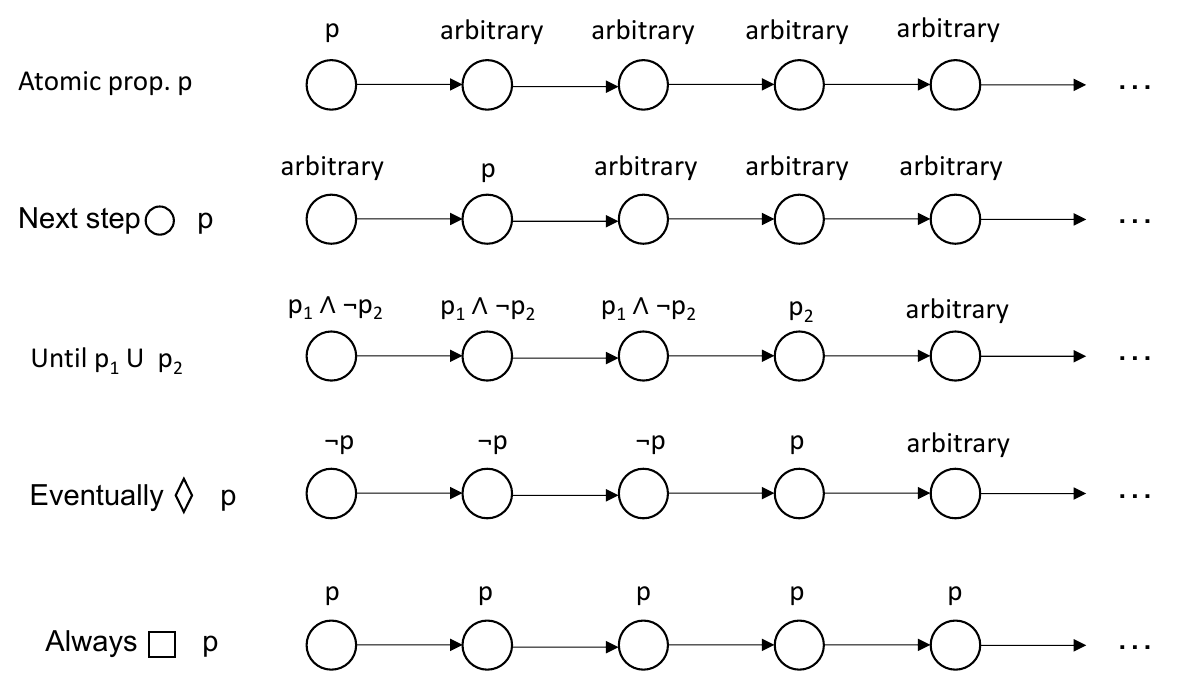}
    \centering
    \caption{Sketch of semantics of temporal modalities}
\end{figure}

Now we go back to the example ``go down that ladder and walk right on the conveyor belt but do not fall off the conveyor belt''. It can be expressed in LTL as $\eventually (p_1 \land p_2) \land \always \neg p_3$, where $p_1, p_2, p_3$ are the propositions that are true if and only if the corresponding atomic events ``go down that ladder'', ``walk right on the conveyor belt''  , ``fall off the conveyor belt'' are performed. And we say an agent trajectory is consistent with the temporal constraints of the language instruction iff the trajectory satisfies the LTL formula of the instruction. Note that the LTL formulae address only the temporal relationship between events, but not deal with the recognition of events. 

LTL supports non-Markovian constraints, though the setting of the task environment in this work is Markovian. However, we will not go into details of that. 

\section{Raw Results Plots}
\label{sec:appendixplots}
We list all the raw result plots with the format shown in Table~\ref{tab:rawresultformat}:
\begin{table}[H]
\centering
\caption{Structure of the raw result plots} 
\label{tab:rawresultformat}
\begin{tabular}{|p{2.9cm}|p{2.9cm}|p{2.9cm}|p{2.9cm}|p{2.9cm}|} 
\hline
cumulative rewards for the first instruction sentence & cumulative rewards for the second instruction sentence & cumulative rewards for the third instruction sentence & cumulative rewards for the fourth instruction sentence & cumulative rewards for the fifth instruction sentence                                        \\ [28pt]
\hline
Number of cumulative achieved state constraints       & Number of cumulative achieved action constraints       & Number of win over training samples                    & Global training steps                                   & gaming steps (i.e., duration) per training episode                                            \\ [28pt]
\hline
Environmental Rewards per training episode             & Number of policy model update                           & Number of sample episodes                              & intrinsic reward per global update (RND algorithm)      & max probability of an action per global update (measures the confidence level of the policy)  \\[8pt]
\hline
\end{tabular}
\end{table}

\begin{figure}[H]
    \includegraphics[width=0.96\textwidth]{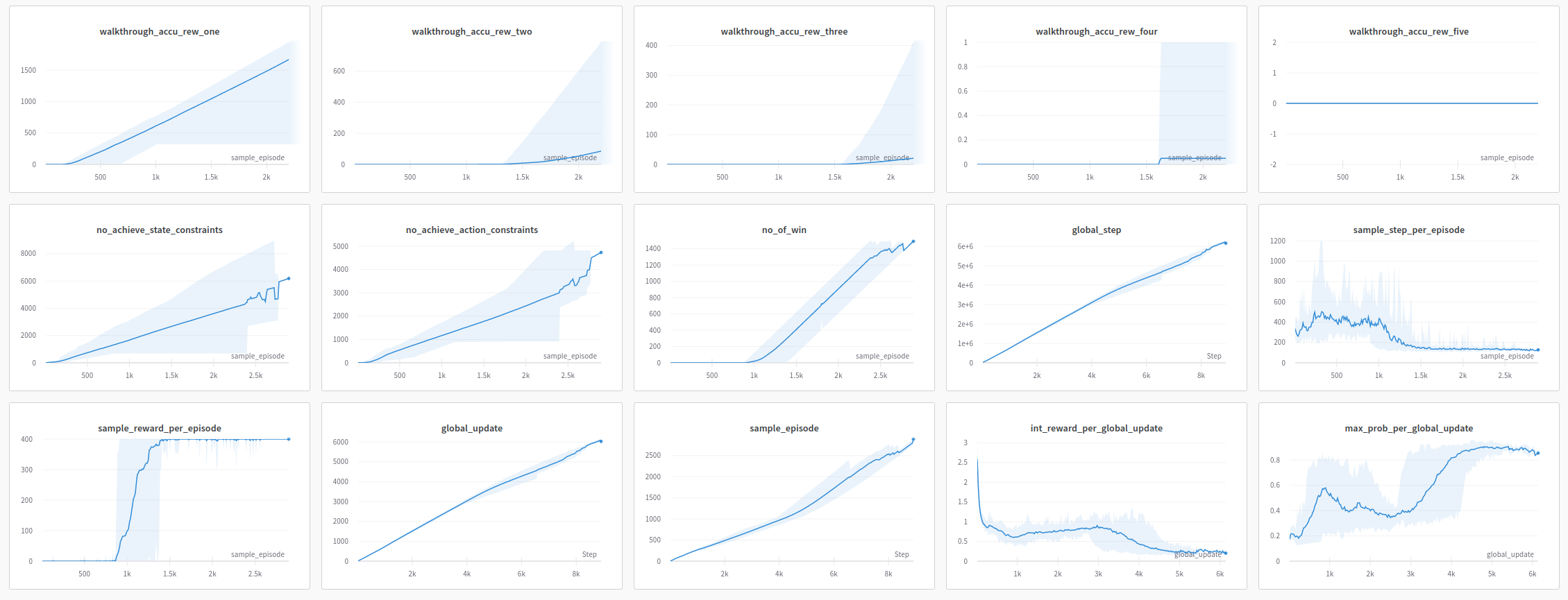}
    \centering
    \caption{LRS Model A raw results plot}
\end{figure}
\begin{figure}[H]
    \includegraphics[width=0.96\textwidth]{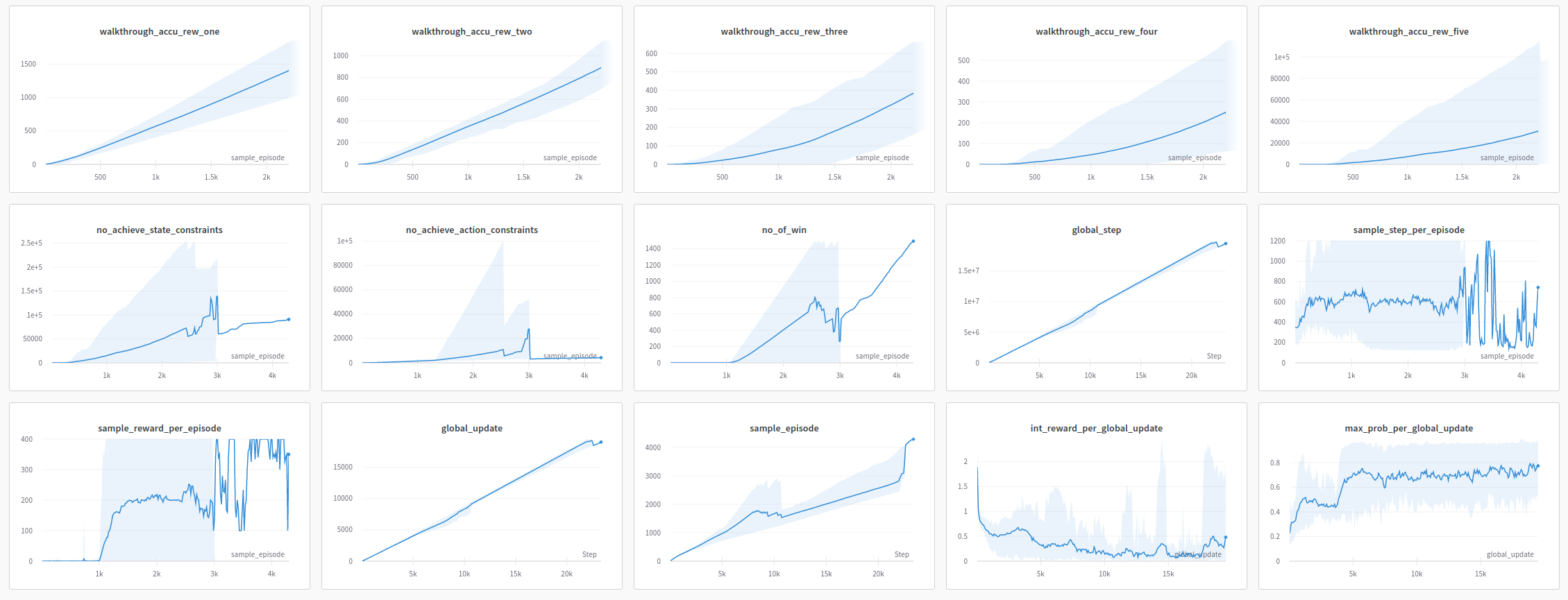}
    \centering
    \caption{LRS Model B raw results plot}
\end{figure}
\begin{figure}[H]
    \includegraphics[width=0.96\textwidth]{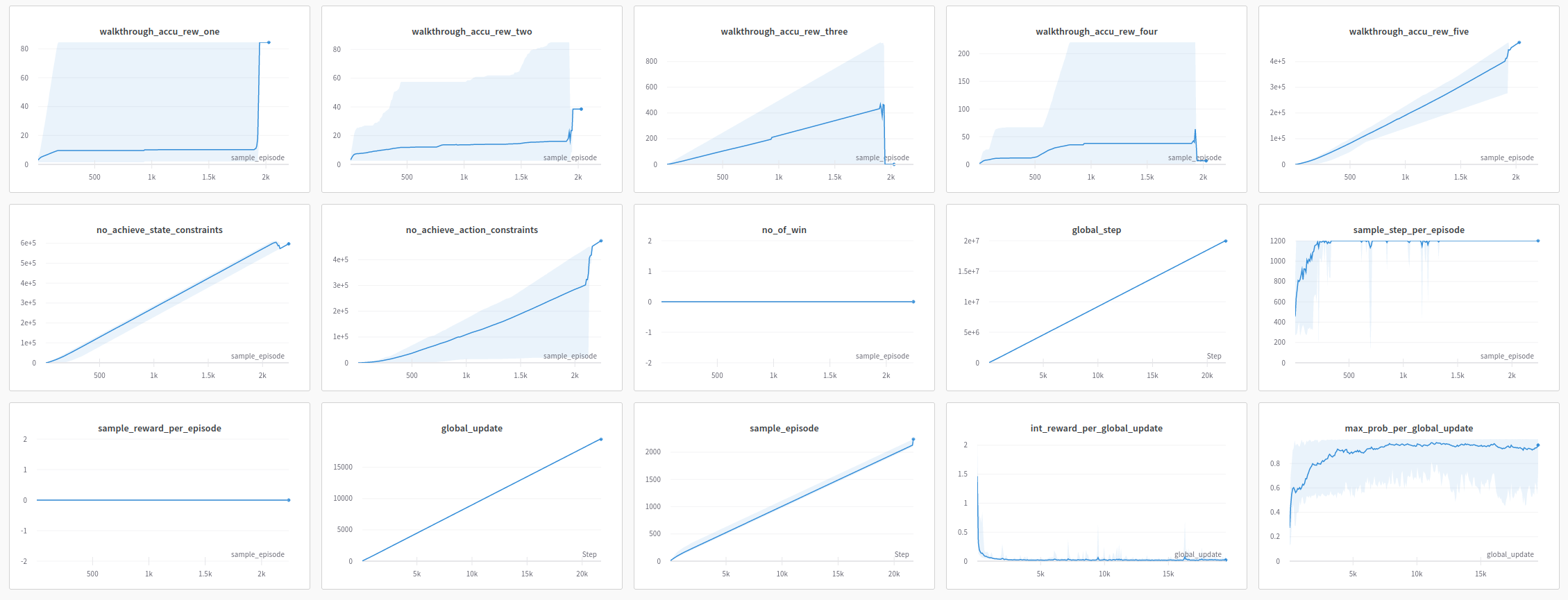}
    \centering
    \caption{LRS Model C raw results plot}
\end{figure}
\begin{figure}[H]
    \includegraphics[width=0.96\textwidth]{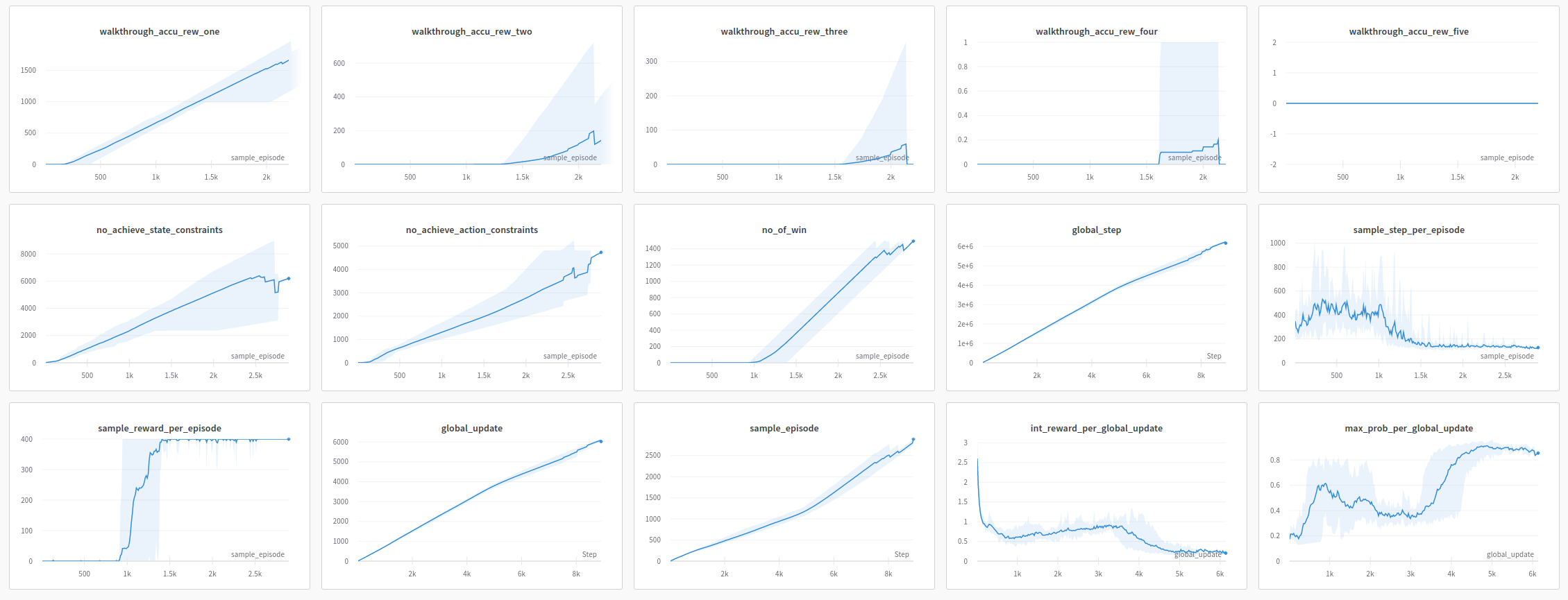}
    \centering
    \caption{LRS Model A with Low Granularity Type 1 raw results plot}
\end{figure}
\begin{figure}[H]
    \includegraphics[width=0.96\textwidth]{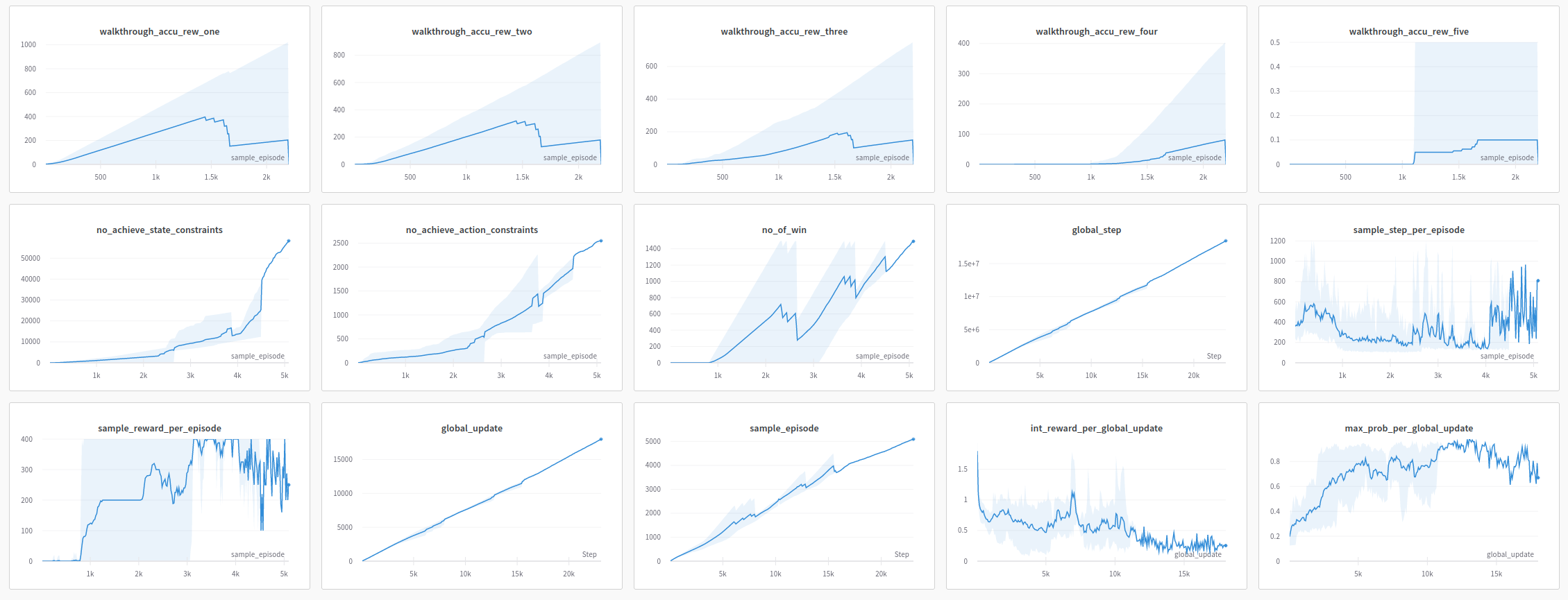}
    \centering
    \caption{LRS Model A with Low Granularity Type 2 raw results plot}
\end{figure}
\begin{figure}[H]
    \includegraphics[width=0.96\textwidth]{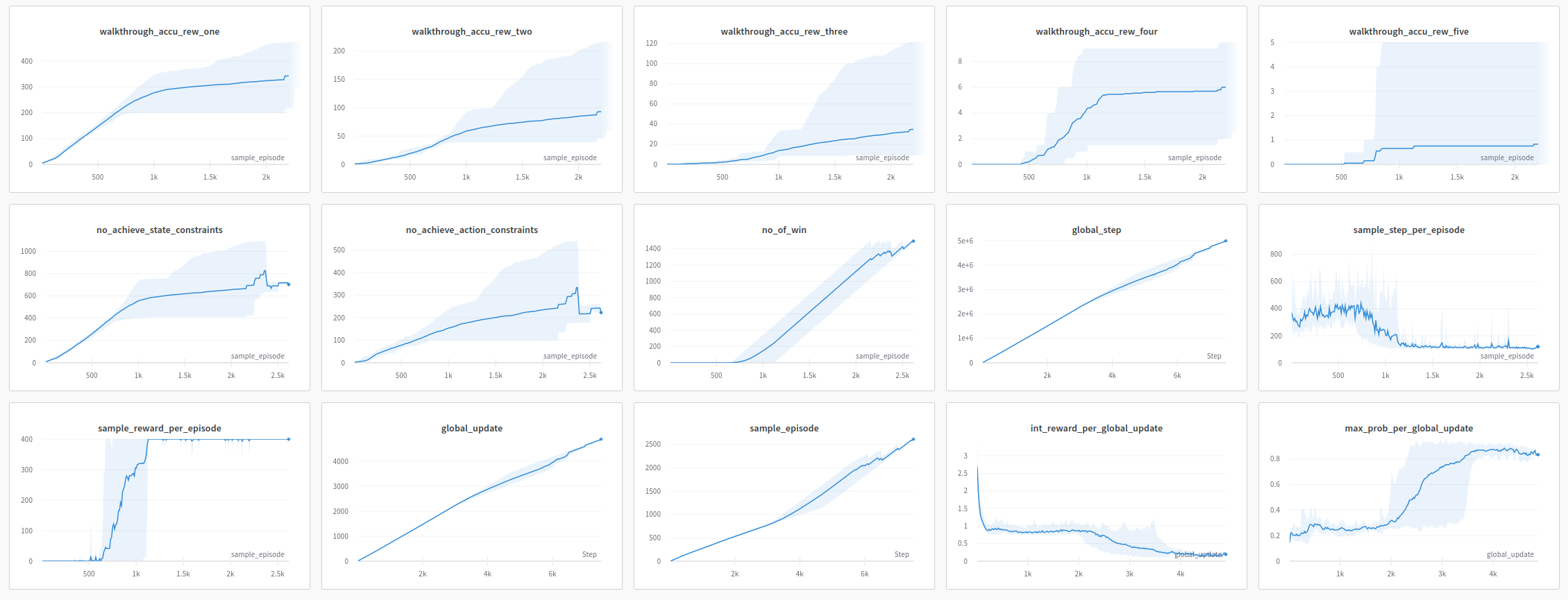}
    \centering
    \caption{Vanilla PPO+RND raw results plot}
\end{figure}

\section{Implementation Details of Non-Simulated LRS Binary Classifier}
\label{sec:appendixadditionalstudies}

Originally this work was targeting to improve the precision score of LRS binary classifier. Although this paper proved that such a mechanism leads to performance deterioration of the instruction following agent, we would like still list some of the techniques we found useful for improving model's precision. Notice that some methods may be specific to Montezuma's Revenge game environment. 

\subsection{Pretraining a visual autoencoder to compress visual observations into dense feature vectors}

Pretraining a visual encoder such that pixel-based visual observations (e.g., $T \times 3 \times 84\times 84$ rgb video clip of length $T$) can be compressed into dense features vectors that capture spatial and temporal information. It helps to reduce the input dimension and thereby speeding up the learning process. Many recent work on RL used such a pretraining method \citep{hafner2020mastering, seo2022masked}. 

\subsection{Soft-discretisation of visual inputs}
\begin{figure}[ht]
    \includegraphics[width=0.4\textwidth]{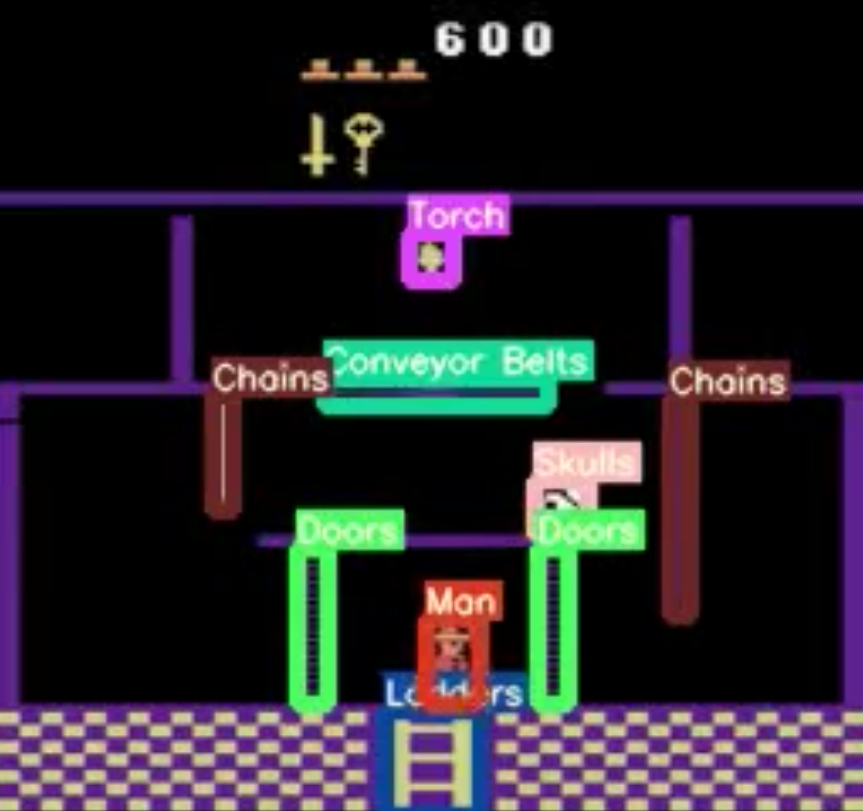}
    \centering
    \caption{An example of having an object detector to annotate the visual input}
    \label{fig:objectdetector}
\end{figure}

With the help of an object detector, the visual perception module of the instruction-following agent can to some extent discretizes the pixel-based visual information to a object-based one (see Figure~\ref{fig:objectdetector}). Given that reasoning process only applies to atomic evidence, and also that physical reasoning in most instruction-following tasks is conducted at the object level, it is argued that the ``discretisation'' process helps to improve the overall performance of the reasoning model. This method has been discussed in recent studies. \citet{hu2019you} suggested that object presentation increased the success rate for Visual Language Navigation tasks. \citet{ding2021attention} also mentioned the performance improvement on Visual Question Answering (VQA) tasks by converting the pixel images into object-based representations first. 

\subsection{Relative offsets to capture the spatial relationships}
Instead of using the original visual observation $obs_t$ as input at time $t$, one can use the relative offset $obs_t - obs_{t-1}$ in order to emphasise the spatial relationship between objects. This approach is also known as B-PRO from \citet{talvitie2015pairwise}, which has been shown effective in shortening the required training steps. This approach is effective especially for Montezuma's Revenge game because the relative offset operation can effectively cancel out the unimportant background information. It remains unclear that this approach is still useful when the environment background becomes dynamic and noisy.  

\subsection{Phrase corruption technique, a clever way to generate hard negative training samples}
\begin{figure}[ht]
    \includegraphics[width=0.85\textwidth]{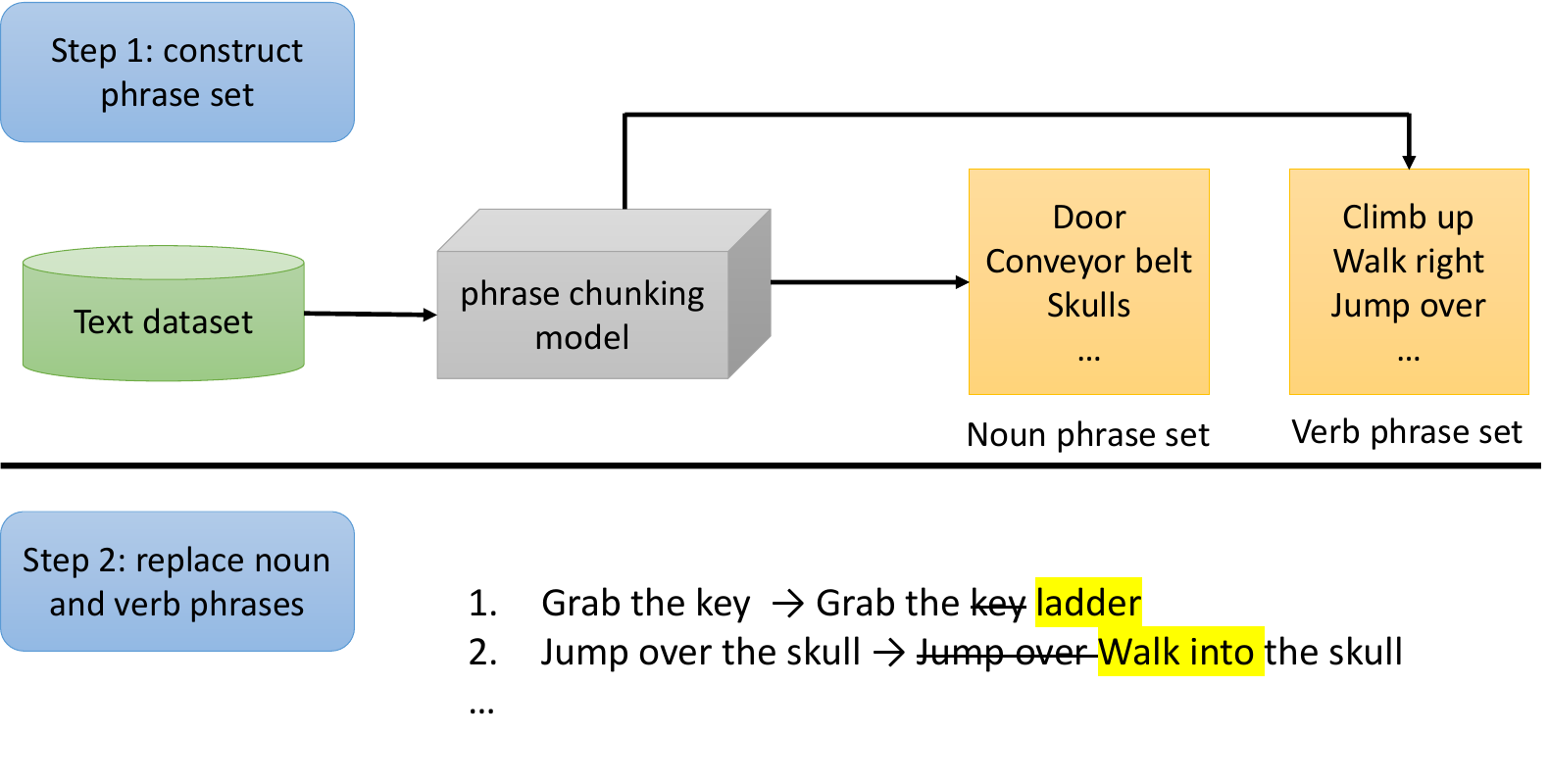}
    \centering
    \caption{An illustration of the ``phrase corruption'' procedure}
    \label{fig:phrasecorruption}
\end{figure}

Training with hard negative samples helps to reduce the false positive errors \citep{xuan2020hard}. However, the trajectory-instruction data pairs for training are all positive pairs and most common techniques such as in-batch negative sampling are not able to generate hard-negatives.
Language instructions are made up of object and action constraints. In light of this, we propose the ``phrase corruption'' technique -- it will ``corrupt'' the object and action constraints by only replacing the action and the object phrases in the language sentences. We consider the new generated instructions as hard negative samples because only the semantic of the instructions are changed while the original sentence structures and the tones are largely maintained. The procedure is as follows -- first, we use a off-the-shelf phase chunking model to chunk the instruction into phrases, meanwhile we extract all the verb and noun phrases from the dataset into two separate phrase set. After that, for each instruction sentence, we randomly select one phrase from the phrase set to replace the original noun and verb phrases (see Figure~\ref{fig:phrasecorruption}). Although this method can help to generate a large number of hard-negative texts, some samples may not logically make sense (e.g., climb down the ladder $\rightarrow$ \st{climb down} \hl{kill} the ladder). 

\subsection{Implementation details during policy training}

In the policy training process, we manually partition the language instructions into atomic sentences. For each policy update iteration, the agent's past trajectory is converted into embedding vectors and fuse together with the currently focused atomic sentence. We assume the agent has fulfilled the current instruction when the matching score exceeds a manually defined threshold, causing the activation of the next atomic sentence. 

\subsection{Overall results on different ablation settings}
We trained a LRS binary classifier on an annotated trajectory dataset from \citet{goyal2019using}. It consists of 5000 annotated human player replay clips with 3 second each. Annotations are obtained by using Amazon Mechanical Turk and are made up of one single sentence about what is happening in the 3-second demonstration \footnote{The annotation dataset can be downloaded from \url{http://www.cs.utexas.edu/~pgoyal/atari-lang.zip}}. The testing dataset is made by ourselves which follows the same patterns as the Goyal et al.'s. The performance of the LRS binary classifier with different ablation settings is shown in Table~\ref{tab:binaryclsablation}.


\begin{table}[H]
\centering
\caption{Evaluation of the metrics on different ablation settings for LRS binary classifier}
\label{tab:binaryclsablation}
\begin{tabular}{|p{1.8cm}|p{1.8cm}|p{1.8cm}|p{1.8cm}||c|c|c|} 
\hline
\multicolumn{4}{|c||}{\textbf{Ablation Settings}}                  & \multirow{2}{*}{Precision}                    & \multirow{2}{*}{Recall} & \multirow{2}{*}{Bal Acc}  \\ 
\cline{1-4}
Pretraining & Discretisation & Relative-Offset & Phrase Corruption &                                               &                         &                                \\ 
\hline
\xmark        & \xmark           &                 &                   & 40.00\%                                       & 40.00\%                 & 55.00\%                        \\ 
\hline
\xmark        & \xmark           & \xmark            &                   & 50.00                                         & 40.00\%                 & 60.00\%                        \\ 
\hline
\xmark        & \xmark           & \xmark            & \xmark              & 66.67\%                                       & 80.00\%                 & 70.00\%                        \\
\hline
\end{tabular}
\end{table}

\end{document}